\documentclass{article}
\usepackage{ijcai25}

\usepackage{times}
\usepackage{soul}
\usepackage{url}
\usepackage[hidelinks]{hyperref}
\usepackage[utf8]{inputenc}
\usepackage[small]{caption}
\usepackage{graphicx}
\usepackage{amsmath}
\usepackage{amsthm}
\usepackage{booktabs}
\usepackage{algorithm}
\usepackage{algorithmic}
\usepackage[switch]{lineno}

\usepackage{amssymb}
\usepackage{amsthm}
\usepackage{array}
\usepackage{enumitem}
\DeclareMathOperator*{\argmin}{arg\,min}
\usepackage{graphicx}
\usepackage{listings}
\usepackage{makecell}
\usepackage{siunitx}
\usepackage{subcaption}
\usepackage{multirow}
\usepackage{tabularx}
\newcommand{\minitab}[2][l]{\begin{tabular}{#1}#2\end{tabular}}
\newcolumntype{P}[1]{>{\centering\arraybackslash}p{#1}}

\newtheorem{example}{Example}

\title{Scalable Multi-Stage Influence Function for Large Language Models via Eigenvalue-Corrected Kronecker-Factored Parameterization}

\newcommand*\samethanks[1][\value{footnote}]{\footnotemark[#1]}

\author{
Yuntai Bao$^1$\and
Xuhong Zhang$^1$\thanks{Corresponding author} \and
Tianyu Du$^1$\samethanks \and
Xinkui Zhao$^1$\and\\
Jiang Zong$^2$\and
Hao Peng$^3$\And
Jianwei Yin$^1$\\
\affiliations
$^1$Zhejiang University \\
$^2$Universal Identification Technology (Hangzhou) Co.,Ltd. \\
$^3$Zhejiang Normal University\\
\emails
\{yuntaibao, zhangxuhong, zjradty, zhaoxinkui\}@zju.edu.cn,\\
zongj@kingflying.cn,
hpeng@zjnu.edu.cn,
zjuyjw@cs.zju.edu.cn
}

\begin{document}

\maketitle

\begin{abstract}
Pre-trained large language models (LLMs) are commonly fine-tuned to adapt to downstream tasks. Since the majority of knowledge is acquired during pre-training, attributing the predictions of fine-tuned LLMs to their pre-training data may provide valuable insights. Influence functions have been proposed as a means to explain model predictions based on training data. However, existing approaches fail to compute ``multi-stage'' influence and lack scalability to billion-scale LLMs.

In this paper, we propose the multi-stage influence function to attribute the downstream predictions of fine-tuned LLMs to pre-training data under the full-parameter fine-tuning paradigm. To enhance the efficiency and practicality of our multi-stage influence function, we leverage Eigenvalue-corrected Kronecker-Factored (EK-FAC) parameterization for efficient approximation.
Empirical results validate the superior scalability of EK-FAC approximation and the effectiveness of our multi-stage influence function. Additionally, case studies on a real-world LLM, dolly-v2-3b, demonstrate its interpretive power, with exemplars illustrating insights provided by multi-stage influence estimates.
\end{abstract}

\section{Introduction}

Understanding the relationship between training data and model behavior is essential for building trustworthy machine learning systems. For example, attributing a model's answer in a closed-book question-answering system to a specific Wikipedia article can enhance user trust.
Training data attribution (TDA) techniques quantify the contributions of training instances to model behaviors by addressing a counterfactual question: how would the model's behavior change if an example were removed from the training set? Originating from robust statistics~\cite{hampel1974influence} and introduced into deep learning by Koh and Liang~\shortcite{koh2017understanding}, influence function (IF) provides an end-to-end, scalar-valued interpretation of a model's high-level behavior.

While several works on IFs have explored their conceptual framework and applications, they often overlook the scalability of these methods to large-scale neural networks trained on massive datasets. Furthermore, prior analyses predominantly focus on classical architectures, such as feed-forward networks, rather than modern architectures like transformers.
Findings of Zhou et al.~\shortcite{zhou2024lima} indicate that most knowledge in large language models (LLMs) is acquired during pre-training. Consequently, explaining the predictions of fine-tuned models necessitates tracing influence back to the pre-training dataset rather than the fine-tuning dataset. Although Chen et al.~\shortcite{chen2020multi} introduced ``multi-stage'' IFs to address this need, their analysis of NLP models is limited to frozen encoders stacked with linear classifiers, failing to accommodate the full-parameter tuning paradigm prevalent in LLMs.

In this work, we propose an IF-based method to estimate the contribution of pre-training data to the predictions of fine-tuned models, scaling efficiently to LLMs with billions of parameters. Our approach addresses two key challenges.

First, the original IF framework~\cite{koh2017understanding} does not accommodate multi-stage influence when fine-tuning tasks require substantial modifications to the model, such as replacing the unembedding layer. Inspired by Chen et al.~\shortcite{chen2020multi}, we extend IFs to the multi-stage paradigm (``pre-train then fine-tune''), enabling attribution of downstream predictions of LLMs to pre-training examples.

Second, scaling IFs to LLMs involves overcoming computational bottlenecks related to inverse Hessian-Vector Product (iHVP) and training gradients.
For iHVPs, we adopt the Eigenvalue-corrected Kronecker-factored Approximate Curvature (EK-FAC) parameterization~\cite{george2018fast}, as suggested by Grosse et al.~\shortcite{grosse2023studying}.
To address the latter bottleneck, we leverage semantic-similarity-based heuristics to narrow the candidate training samples, avoiding iterations over the entire dataset.

We conduct extensive experiments to evaluate our method. For general influence estimation, we demonstrate the superior scalability of EK-FAC approximations compared to various TDA methods. For our multi-stage IF, we evaluate it on our fact-tracing benchmark and show that it outperforms the single-stage IF~\cite{grosse2023studying}.
Additionally, we analyze the contributions of MLP and multi-head attention (MHA) parameters to influence estimation, finding that MLP parameters play a proportionally greater role. This insight suggests a practical trade-off, allowing analysis to focus on MLP parameters in large models.
Finally, we apply our multi-stage IF on a publicly available instruction-tuned LLM, dolly-v2-3b~\cite{DatabricksBlog2023DollyV2}, qualitatively explaining its generations based on pre-training data.

In summary, we propose a general framework for efficiently estimating multi-stage influence with the help of EK-FAC parameterization and semantic-similarity-based candidate selection. Our results provide practical insights into resolving influence estimation trade-offs. Furthermore, we demonstrate how TDA approaches can be applied to calibrate the trustworthiness of generative AI systems.

\section{Related Work}
\paragraph{Training data attribution.}
Training data attribution (TDA) techniques explain a model's predictions by quantifying the contribution of training data. As noted by Hammoudeh and Lowd~\shortcite{hammoudeh2024training}, TDA methods can be broadly categorized into retraining-based and gradient-based approaches.
\textit{Retraining-based} methods estimate the influence of individual examples by retraining models on random subsets of the training dataset. However, these methods incur high computational costs due to the need for multiple retraining rounds, rendering them impractical for large-scale models and datasets.
\textit{Gradient-based} methods, which infer training data influence using gradients, are further divided into dynamic and static approaches. Dynamic estimators, such as TracIn~\cite{pruthi2020estimating}, assess influence by analyzing gradients from intermediate model snapshots captured during training. In contrast, static estimators, including IFs~\cite{koh2017understanding} and representer point selection~\cite{yeh2018representer}, rely solely on the final model parameters to compute influence.

\paragraph{Influence functions.}
Despite their utility, IFs exhibit several limitations.
First, in terms of \textit{model architecture}, most existing studies focus on traditional architectures such as feed-forward and recurrent networks, with limited exploration of modern architectures like transformers. A recent study by Grosse et al.~\shortcite{grosse2023studying} extended IF analysis to transformer-based LLMs using EK-FAC approximation. In this work, we also investigate transformer language models.

Second, regarding \textit{scalability}, most research adopts a ``matrix-free'' approach to avoid the computational costs of explicitly and inverting the Hessian for large models, but with limited success. One prominent strategy involves parametric approximations of the Hessian to enable efficient inverse Hessian computations~\cite{schioppa2022scaling,grosse2023studying}. Another approach uses iterative stochastic approximation methods, such as LiSSA and Conjugate Gradient, to approximate iHVPs, as introduced by Koh and Liang~\shortcite{koh2017understanding}. However, our experiments show that these methods fail to yield usable influence estimates within a practical timeframe.

Third, on the \textit{training paradigm}, the original IF proposed by Koh and Liang~\shortcite{koh2017understanding} is unsuitable for analyzing the influence of pre-training data on a fine-tuned model when it has a different output domain. While Grosse et al.~\shortcite{grosse2023studying} proposed an efficient method for single-stage training scenarios, it does not address multi-stage paradigms. Chen et al.~\shortcite{chen2020multi} introduced multi-stage IF as a generalization of the original IF, but their work is limited to the ELMo~\cite{peters-etal-2018-deep} architecture and fine-tuning scenarios where a classification head is added to a frozen pre-trained encoder. In contrast, we focus on the popular full-parameter fine-tuning paradigm.

\section{Background and Notations}
\subsection{Transformer Architecture}
The focus of our work is the transformer language model~\cite{vaswani2017attention}, which starts with a token embedding, followed by a series of $L$ residual blocks, and ends with a token unembeddings~\cite{elhage2021mathematical} (layer normalization is omitted for brevity). For a sequence $t$ with $n$ tokens, the initial embeddings are $\mathbf{x}_0=\text{Embed}(t)\in \mathbb{R}^{n\times d}$, hence the start of the residual stream.

At each residual block, the residual stream is first processed by the MHA module: $\tilde{\mathbf{x}}_l = \mathbf{x}_{l-1} + \sum_{h\in H_l}h(\mathbf{x}_{l-1})$, where $H_l$ is the set of attention heads and $\mathbf{x}_{l-1}$ is the input for the $l$-th ($1 \leq l \leq L$) residual block.
The attention pattern for head $h$ is obtained via attention mechanism: $\mathbf{r}^h = \text{Attention}\left(\mathbf{q}^h, \mathbf{k}^h, \mathbf{v}^h\right)$, where $\mathbf{q}^h=\mathbf{x}_{l-1}\left(\mathbf{W}_Q^h\right)^\top$, $\mathbf{k}^h=\mathbf{x}_{l-1}\left(\mathbf{W}_K^h\right)^\top$, $\mathbf{v}^h=\mathbf{x}_{l-1}\left(\mathbf{W}_V^h\right)^\top$. $\mathbf{W}_Q^h, \mathbf{W}_K^h, \mathbf{W}_V^h \in \mathbb{R}^{n_{\text{context}}^H\times d}$ are weights for query, key and value, respectively. The attention pattern is then written into the residual stream: $h(\mathbf{x}_{l-1})=\mathbf{r}^h \left(\mathbf{W}_O\right)^\top$, $\mathbf{W}_O \in \mathbb{R}^{d \times n_{\text{context}}^H}$.

The MLP module then processes the output of the MHA and writes back to the residual stream: $\mathbf{x}_l = \tilde{\mathbf{x}}_l + \sigma \left(\tilde{\mathbf{x}}_l \left(\mathbf{W}_I^m\right)^\top \right) \left(\mathbf{W}_O^m\right)^\top$, where $\sigma(\cdot)$ is the element-wise nonlinear activation, $\mathbf{W}_I^m \in \mathbb{R}^{n_{\text{context}}^m \times h}$ and $\mathbf{W}_O^m \in \mathbb{R}^{h \times n_{\text{context}}^m}$ are projection weights. After $L$ residual blocks, the unembeddings produce the final logits: $T(t) = \mathbf{x}_L \mathbf{W}_U^\top$.

\subsection{Influence Functions}
In this section, we demonstrate the original single-stage IF~\cite{koh2017understanding}. This formulation quantifies the influence of a training example on both model parameters and a measurement of a query sample.

Consider a training dataset $\mathcal{D}=\{z_i\}_{i=1}^N$, where each example $z_i$ represents a token sequence, and the learning task is self-supervised language modeling. The training objective is to minimize the expectation $\mathcal{L}$ of the loss function $\ell(\cdot)$:
\begin{equation}
\begin{aligned}
\boldsymbol{\theta}^*
= \argmin_{\boldsymbol{\theta}} \mathcal{L}(\boldsymbol{\theta}, \mathcal{D}) = \argmin_{\boldsymbol{\theta}} \frac{1}{N}\sum _{i=1}^{N} \ell(z_i, \boldsymbol{\theta}).
\end{aligned}
\end{equation}

The influence of a training sample $z$ on the optimal parameters $\boldsymbol{\theta}^*$ is $\mathcal{I}_{\boldsymbol{\theta}^*}(z_m)$, while the influence on a query $z_q$ is mediated via a differentiable measure, $m(z_q,\boldsymbol{\theta})$, e.g., autoregressive cross-entropy loss of $z_q$. The influence of $z$ with respect to $\boldsymbol{\theta}^*$ can thus be expressed as:
\begin{equation}
\begin{aligned}
    \mathcal{I}_{m}(z, z_q) &= \nabla_{\boldsymbol{\theta}}m(z_q, \boldsymbol{\theta}^*)^\top \mathcal{I}_{\boldsymbol{\theta}^*}(z) \\
    &= \nabla_{\boldsymbol{\theta}}m(z_q, \boldsymbol{\theta}^*)^\top \mathbf{H}_{\boldsymbol{\theta}^*}^{-1}\nabla_{\boldsymbol{\theta}}\ell(z, \boldsymbol{\theta}^*),
\end{aligned}
\end{equation}
where $\mathbf{H}_{\boldsymbol{\theta}^*}=\nabla_{\boldsymbol{\theta}}^2 \mathcal{L}(\boldsymbol{\theta}^*,\mathcal{D})$.

In practice, the Hessian may be singular when the model is not fully converged. To address this, we follow Bae et al.~\shortcite{bae2022if} and replace the Hessian with a damped generalized Gauss-Newton (GGN) matrix~\cite{schraudolph2002fast,martens2020new} to ensure positive-definiteness. This modification enables the use of the final model parameters in place of strictly converged parameters $\boldsymbol{\theta}^*$. Additionally, we use the cross-entropy loss to ensure that the loss function is convex with respect to model outputs.

\subsection{EK-FAC for Feed-Forward Networks}\label{sec:ekfac_mlp}
Naive computation of the GGN ($\mathbf{G}$) or its inverse is computationally prohibitive. To address this, EK-FAC~\cite{george2018fast} was proposed as an efficient approximation method for computing iHVPs. This approach was subsequently adopted by Grosse et al.~\shortcite{grosse2023studying} to estimate single-stage influence for LLMs.
Structurally, the GGN is approximated as a block-diagonal matrix. Each diagonal block is then approximated using EK-FAC parameterization.

Consider a feed-forward neural network with $L$ layers interleaved with nonlinear activations. Let the parameters be $\mathbf{W}=(\mathbf{W}_1, \mathbf{W}_2, ..., \mathbf{W}_L)$ and $\boldsymbol{\theta}=(\boldsymbol{\theta}_1^\top,...,\boldsymbol{\theta}_L^\top)^\top$, where $\boldsymbol{\theta}_l = \text{vec}(\mathbf{W}_l)$ denotes the vectorized form of $\mathbf{W}_l (l=1,...,L)$. The GGN for this network is expressed as:
\begin{equation}
\begin{aligned}
    \mathbf{G} = \underset{\begin{subarray}{c}
    (x_n,y_n)\sim \hat{Q}_x\\
    y\sim p(y|f_{\boldsymbol{\theta}}(x_n))
    \end{subarray}}{\mathbb{E}} \left[ \mathcal{D}\boldsymbol{\theta} \mathcal{D}\boldsymbol{\theta} ^\top \right] 
    =\left[ \mathbf{G}_{i,j} \right]_{1 \leq i,j \leq L},
\end{aligned}
\end{equation}
where $\hat{Q}_x$ is the training dataset, $\mathcal{D}\boldsymbol{\theta} = \nabla_{\boldsymbol{\theta}}\log p(y|f_{\boldsymbol{\theta}}(x_n),\boldsymbol{\theta})$, and $\mathbf{G}_{i,j}=\mathbb{E}\left[\mathcal{D}\boldsymbol{\theta}_i \mathcal{D}\boldsymbol{\theta}_j^\top\right]$ is a block of the GGN. Note that the label $y$ is sampled from the model's predictive distribution rather than the training label.

Approximating the GGN as block-diagonal retains only the diagonal blocks: $\mathbf{G} \approx \tilde{\mathbf{G}} = \text{diag}(\mathbf{G}_{1,1}, ..., \mathbf{G}_{L,L})$. Given a vector $\boldsymbol{v}$ with the same dimensions as $\boldsymbol{\theta}$, the damped iHVP can be computed as $\left(\tilde{\mathbf{G}}+\lambda \mathbf{I}\right)^{-1}\mathbf{v} = \text{diag}\left(\left(\mathbf{G}_{1,1}+\lambda \mathbf{I}\right)^{-1}\mathbf{v}_1, ..., \left(\mathbf{G}_{L,L}+\lambda \mathbf{I}\right)^{-1}\mathbf{v}_L \right)$, where $\mathbf{v}_l$ is the slice of $\mathbf{v}$ corresponding to $\boldsymbol{\theta}_l$.

For each block $\mathbf{G}_{l,l}$, EK-FAC provides a further approximation. Let the inputs and outputs of the $l$-th layer be $\mathbf{a}_{l-1}$ and $\mathbf{s}_l$, respectively. The gradient $\mathcal{D}\boldsymbol{\theta}_l$ is expressed as $\mathbf{a}_{l-1} \otimes \mathcal{D}\mathbf{s}_l$, where $\otimes$ denotes Kronecker product. $\mathbf{G}_{l,l}$ can thus be approximated using K-FAC as:
\begin{equation}
    \begin{aligned}
        \mathbf{G}_{l,l} &= \mathbb{E}\left[ \left(\mathbf{a}_{l-1} \mathbf{a}_{l-1}^\top\right) \otimes \left(\mathcal{D}\mathbf{s}_l \mathcal{D}\mathbf{s}_l^\top\right) \right] \\
        &\approx \mathbb{E} \left[ \mathbf{a}_{l-1} \mathbf{a}_{l-1}^\top \right] \otimes \mathbb{E} \left[ \mathcal{D}\mathbf{s}_{l} \mathcal{D}\mathbf{s}_{l}^\top \right] \\
        &=  \mathbf{A}_{l-1,l-1} \otimes  \mathbf{S}_{l,l} = \tilde{\mathbf{G}}_{l,l}.
    \end{aligned}
\end{equation}

EK-FAC improves upon K-FAC by incorporating the diagonal variance in the eigenbasis of $\mathbf{A}_{l-1,l-1}$ and $\mathbf{S}_{l,l}$. Denoting these matrices as $\mathbf{A}$ and $\mathbf{S}$ for brevity, their eigendecompsition yields:
\begin{equation}
    \begin{aligned}
        \tilde{\mathbf{G}}_{l,l} &= \mathbf{A} \otimes \mathbf{S} \\
        &= (\mathbf{Q}_{\mathbf{A}} \otimes \mathbf{Q}_{\mathbf{S}}) (\mathbf{\Lambda}_{\mathbf{A}} \otimes \mathbf{\Lambda}_{\mathbf{S}}) (\mathbf{Q}_{\mathbf{A}} \otimes \mathbf{Q}_{\mathbf{S}})^\top,
    \end{aligned}
\end{equation}
where $\mathbf{Q}_{\mathbf{A}}$ and $\mathbf{Q}_{\mathbf{S}}$ are eigenvectors, and $\mathbf{\Lambda}_{\mathbf{A}}$ and $\mathbf{\Lambda}_{\mathbf{S}}$ are diagonal matrices of eigenvalues. To account for the diagonal variance, the middle factor is replaced by a new diagonal matrix $\mathbf{\Lambda}$ with its diagonal entries as follows:
\begin{equation}
    \begin{aligned}
        \mathbf{\Lambda}_{ii} &= \mathbb{E}\left[\left( \left(\mathbf{Q_A}\otimes\mathbf{Q_S}\right)^\top \mathcal{D}\boldsymbol{\theta}_l \right)_i^2\right] \\
    &= \mathbb{E}\left[ \left( \text{vec}\left( \mathbf{Q_S}^\top \mathcal{D}\mathbf{W}_l \mathbf{Q_A} \right) \right)_i^2 \right].
    \end{aligned}
\end{equation}

Finally, the damped iHVP for the $l$-th block is computed as:
\begin{equation}
\begin{alignedat}{2}
    ({\mathbf{G}}_{l,l}+\lambda \mathbf{I})^{-1}\mathbf{v}_l & \approx && (\tilde{\mathbf{G}}_{l,l}+\lambda \mathbf{I})^{-1}\mathbf{v}_l \\
    & \approx && \left(\mathbf{Q_A}\otimes\mathbf{Q_S}\right) \mathbf{\Lambda}_{\lambda}^{-1}  \left(\mathbf{Q_A}\otimes\mathbf{Q_S}\right)^\top \mathbf{v}_l \\
    & = &&\text{vec} \Bigg(\mathbf{Q_S}\bigg[ \left(\mathbf{Q_S}^\top \bar{\mathbf{V}_l} \mathbf{Q_A}\right) \oslash \\
    & &&\text{unvec}\left(\text{diag}^{-1} \left(\mathbf{\Lambda}_{\lambda}\right)\right)\bigg]\mathbf{Q_A}^\top\Bigg),
\end{alignedat}
\end{equation}
where $\mathbf{\Lambda}_{\lambda} = \mathbf{\Lambda} + \lambda \mathbf{I} (\lambda>0)$, $\oslash$ denotes element-wise division, $\text{diag}^{-1}(\cdot)$ extracts the diagonal elements of a matrix into a vector, and $\text{unvec}(\cdot)$ converts a vector into a matrix.

\section{Method} \label{sec:method_body}
In this section, we present the design of our multi-stage IF, which attributes the predictions of a fine-tuned LLM to its pre-training data. Additionally, we describe the use of approximation techniques to make the multi-stage IF computationally tractable for LLMs, addressing the trade-off between efficiency and effectiveness.

The objective of the multi-stage IF is to quantify the influence of a pre-training sample $z$ on a test-time query sample $x_t$. The estimation process consists of two primary steps: candidate selection (Section \ref{sec:candidate_filtering}) and influence computation (Section \ref{sec:ekfac_formulation}).
First, given a query, we filter the pre-training dataset using similarity heuristics for a much smaller subset of training examples as candidates for influence estimation. This step ensures that the selection is focused on training examples that are semantically relevant to the query.
In the second step , we compute the influence of the selected candidates on the query, producing a series of influence scores.

\subsection{Multi-Stage Influence Function} \label{sec:multi_stage_influence_function}
Before formulating the multi-stage IF, we first review the ``pre-train then fine-tune'' paradigm. During pre-training, all model parameters are randomly initialized and subsequently updated to fit the distribution of a large-scale corpus.
\begin{equation}
\begin{aligned}
\boldsymbol{\theta}^{\text{pt}} = \argmin_{\boldsymbol{\theta}} \mathcal{L}_{\text{pt}}(\boldsymbol{\theta}) = \argmin_{\boldsymbol{\theta}} \frac{1}{N} \sum_{i=1}^{N} \ell_{\text{pt}}(z_i, \boldsymbol{\theta}),
\end{aligned}
\end{equation}
where $\ell_{\text{pt}}(\cdot)$ is the pre-training loss.

During fine-tuning, the model parameters are initialized with $\boldsymbol{\theta}_{\text{pt}}$ and subsequently optimized on a fine-tuning dataset under the cost function $\ell_{\text{ft}}(\cdot)$.
\begin{equation}
\begin{aligned}
\boldsymbol{\theta}^{\text{ft}} = \argmin_{\boldsymbol{\theta}} \mathcal{L}_{\text{ft}}(\boldsymbol{\theta}) = \argmin_{\boldsymbol{\theta}} \frac{1}{M}\sum_{i=1}^{M} \ell_{\text{ft}}(x_i, \boldsymbol{\theta}).
\end{aligned}
\end{equation}

A naive approach to derive multi-stage influence is as follows:
\begin{equation}
\begin{aligned}
    \mathcal{I}_{m}(z, x) = \nabla_{\boldsymbol{\theta}} m(x, \boldsymbol{\theta}^{\text{ft}})^\top \mathbf{H}_{\boldsymbol{\theta}^{\text{ft}}}^{-1} \nabla_{\boldsymbol{\theta}} \ell_{\text{pt}}(z, \boldsymbol{\theta}^{\text{ft}}).
\end{aligned}
\end{equation}
This formulation is valid as long as the fine-tuned model shares the same output domain as the pre-trained model. For example, it works for pre-trained language models fine-tuned to follow instructions. However, discrepancies may arise when the final unembedding layer is replaced during fine-tuning. For instance, pre-training typically involves the language modeling task with outputs as logits over a large vocabulary, $f_{\boldsymbol{\theta}^{\text{pt}}}(z) \in \mathbb{R}^{|\mathcal{V}|}$, whereas fine-tuning may adapt the model for binary sequence classification, resulting in outputs $f_{\boldsymbol{\theta}^{\text{ft}}}(x) \in \mathbb{R}^{2}$. Consequently, the gradient $\nabla_{\boldsymbol{\theta}} \ell_{\text{pt}}(z, \boldsymbol{\theta}^{\text{ft}})$ is formally invalid for the fine-tuned model.

To address this, we propose making the pre-training gradient accessible to the fine-tuned model by establishing a connection between their parameter spaces. Fine-tuning implicitly assumes that the model retains its general capabilities after adaptation. Thus it is reasonable to assume that the fine-tuned parameters do not deviate significantly from pre-trained ones. Under this assumption, fine-tuning can be viewed as minimizing empirical risk on the fine-tuning task in the vicinity of the pre-trained parameters in the parameter space.

Inspired by Chen et al.~\shortcite{chen2020multi}, we instantiate a quantifiable connection between the pre-trained model and its fine-tuned successor by introducing an additional proximity constraint to the fine-tuning objective in a post-hoc manner. Specifically, we define the proximity as the Euclidean distance between fine-tuned and pre-trained parameters:
\begin{equation}
\begin{aligned}
\boldsymbol{\theta}^{\text{ft}} = \argmin_{\boldsymbol{\theta}} \mathcal{L}_{\text{ft}}(\boldsymbol{\theta}) + \frac{\alpha}{2} || \boldsymbol{\theta}-\boldsymbol{\theta}^{\text{pt}} ||_2^2,
\end{aligned}
\end{equation}
where $\alpha \in \mathbb{R}^+$ is a hyperparameter and $||\cdot||_2$ is L2 norm.

Based on this reformulated objective above, the influence of a pre-training example $z$ on a test instance $x_t$ under the measurement $m(\cdot)$ is given by (full proof in Appendix):
\begin{equation} \label{eq:multi_stage_influence_function_body}
\begin{aligned}
    \mathcal{I}_m(z,x_t) = &\nabla_{\boldsymbol{\theta}} m\left(x, \boldsymbol{\theta}^{\text{ft}}\right)^\top 
                  \left(\nabla_{\boldsymbol{\theta}}^2 \mathcal{L}_{\text{ft}}\left(\boldsymbol{\theta}^{\text{ft}}\right) + \alpha \mathbf{I}\right)^{-1} \\ 
                &\left(\nabla_{\boldsymbol{\theta}}^2 \mathcal{L}_{\text{pt}}\left(\boldsymbol{\theta}^{\text{pt}}\right)\right)^{-1} \nabla_{\boldsymbol{\theta}} \ell_{\text{pt}}\left(z, \boldsymbol{\theta}^{\text{pt}}\right).
\end{aligned}
\end{equation}

\subsection{Practical Implementation for LLMs} \label{sec:ekfac_formulation}
A straightforward approach to implementing IFs involves calculating $\mathbf{H}_{\boldsymbol{\theta}}$ and $\mathbf{H}_{\boldsymbol{\theta}}^{-1}$, followed by computing iHVPs via $\mathbf{H}_{\boldsymbol{\theta}}^{-1}\mathbf{v}$. However, for a model with $p$ parameters and $N$ training samples, $\mathcal{O}(Np^2 + p^3)$ operations are required~\cite{koh2017understanding}, which is computationally prohibitive when $p$ and $N$ are large. To address this, following Grosse et al.~\shortcite{grosse2023studying}, who approximated single-stage IF with EK-FAC, we adopt a similar approach to approximate the multi-stage IF as described in Section \ref{sec:ekfac_mlp}.

First, we replace Hessians with damped GGNs: $\nabla_{\boldsymbol{\theta}}^2 \mathcal{L}_{\text{ft}}\left(\boldsymbol{\theta}^{\text{ft}}\right) \approx \mathbf{G}_{\text{ft}}+\lambda_{\text{ft}}\mathbf{I}$ and $\nabla_{\boldsymbol{\theta}}^2 \mathcal{L}_{\text{pt}}\left(\boldsymbol{\theta}^{\text{pt}}\right) \approx\mathbf{G}_{\text{pt}}+\lambda_{\text{pt}}\mathbf{I}$. For $\mathbf{G}_{\text{ft}}$, the term $\alpha$ in Equation (\ref{eq:multi_stage_influence_function_body}) is absorbed into the damping term $\lambda_{\text{ft}}$. Consequently, the multi-stage IF (Equation (\ref{eq:multi_stage_influence_function_body})) can be interpreted as the inner product of two preconditioned gradients: $\left(\mathbf{G}_{\text{pt}}+\lambda_{\text{pt}}\mathbf{I}\right)^{-1} \nabla_{\boldsymbol{\theta}} \ell_{\text{pt}}\left(z, \boldsymbol{\theta}^{\text{pt}}\right)$ and $\left(\mathbf{G}_{\text{ft}}+\lambda_{\text{ft}}\mathbf{I}\right)^{-1} \nabla_{\boldsymbol{\theta}} m\left(x_t, \boldsymbol{\theta}^{\text{ft}}\right)$. In practice, we implement multi-stage IF following this interpretation.

Next, we compute EK-FAC factors for $\mathbf{G}_{\text{pt}}$ and $\mathbf{G}_{\text{ft}}$, focusing on the linear components of the model. The EK-FAC factors are precomputed and stored on disk, to be loaded when needed. These include $\mathbf{W}_I^m$ and $\mathbf{W}_O^m$ of MLP modules, $\mathbf{W}_Q^h$, $\mathbf{W}_K^h$ and $\mathbf{W}_V^h$ in each attention head, and $\mathbf{W}_O$ of MHA modules. We exclude unembedding parameters, as the unembedding layer of $\boldsymbol{\theta}^{\text{pt}}$ and $\boldsymbol{\theta}^{\text{ft}}$ may have different output dimensions. We also exclude layer normalization modules, since their parameter count is maginal and they are usually not considered to encode factual knowledge~\cite{grosse2023studying}.

We focus on autoregressive decoder-only LLMs, the pre-training loss is the cross entropy loss, following Grosse et al.~\shortcite{grosse2023studying}. For a sequence $z$ with $T$ tokens:
\begin{equation}
    \begin{aligned}
        \ell_{\text{pt}}(z, \boldsymbol{\theta}^{\text{pt}}) = -\sum_{i=1}^{T} \log p_{\hat{y}|x}(z_i|z_{<i};\boldsymbol{\theta}^{\text{pt}}),
    \end{aligned}
\end{equation}
where $p_{\hat{y}|x}$ is the pre-trained model's output distribution.

\subsubsection{Computational and Spatial Cost}
The one-time cost of preparing EK-FAC factors is considered an overhead amortized across future influence analyses.
During influence score computation, for a weight matrix with input dimension $d$ and output dimension $p$, the cost of computing an iHVP is $\mathcal{O}(d^2 p + d p^2)$, followed by an inner product between query iHVP and candidate iHVP at $\mathcal{O}(dp)$. The memory and storage overhead arises from storing eigenvectors $\mathbf{Q}$ and the diagonal entries of $\mathbf{\Lambda}$, resulting in an extra spatial cost of $d^2 + p^2 + dp$.

For example, in the GPT-NeoX~\cite{gpt-neox-library} architecture, where $d=p$ for query, key, value, and output linear projection weights in MHA modules, the additional spatial cost is 3 times the size of the original weights. For MLP modules, where $d=4p$ or $d=\frac{1}{4}p$, the additional spatial cost is 5.25 times the size of the original weights.

\subsection{Selecting Candidates for Influence Estimation} \label{sec:candidate_filtering}
To identify a few positively influential training examples for a query, a naive approach would compute the influence of every training example on the query. However, this requires gradient computations across the entire dataset, a cost equivalent to one epoch of training. To address this, we narrow down the candidate set using efficient similarity-based heuristics inspired by Grosse et al.~\shortcite{grosse2023studying}.

While Grosse et al.~\shortcite{grosse2023studying} employed TF-IDF~\cite{ramos2003tfidf}, we adopt an unsupervised K-Nearest Neighbors (KNN) approach based on the embeddings of pre-training documents, similar to the approach of Guo et al.~\shortcite{guo2021fastif}. The embeddings are generated using Sentence Transformers~\cite{reimers-2019-sentence-bert}, and a KNN classifier is constructed over these embeddings.
This choice reflects the intuition that training examples with similar semantics to the query are more interpretable and relevant than those based solely on textual overlap~\cite{karpukhin2020dense}. Moreover, this method aligns with the potential application of multi-stage IFs in identifying pre-training documents that serve as grounding knowledge sources.

\section{Experiments}
This section presents a series of experiments to evaluate our proposed method. First, we assess the scalability of single-stage IFs approximated using EK-FAC compared to various TDA methods in terms of both estimation accuracy and wall-clock runtime on language modeling tasks (Section \ref{sec:scalability_validation_transformer}).
Next, we investigate the validity of the underlying assumption of our multi-stage IF (Section \ref{sec:examine_euclidean_proximity}).
Subsequently, we evaluate the effectiveness of the multi-stage IF on a factual knowledge retrieval task (Section \ref{sec:multi_stage_infl_effectiveness}).
Finally, we showcase a qualitative case study using the multi-stage IF on an instruction-following LLM (Section \ref{sec:qualitative_studies}).
Our code is public at \url{https://github.com/colored-dye/multi_stage_influence_function}.

\subsection{Scalability Validation of EK-FAC}\label{sec:scalability_validation_transformer}
This experiment evaluates the effectiveness and efficiency of EK-FAC parameterization in producing influence estimates. For simplicity, we focus exclusively on single-stage language modeling rather than the ``pre-train then fine-tune'' scenario.

\begin{figure}
    \centering
    \includegraphics[width=0.98\linewidth]{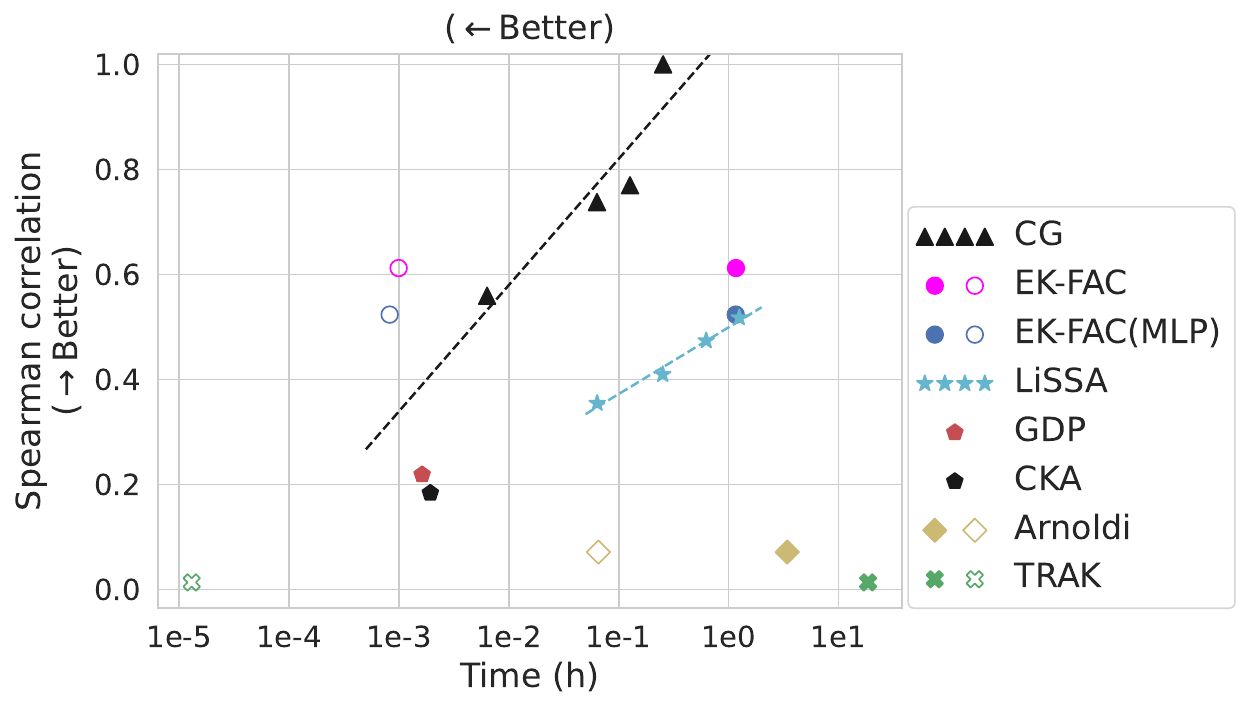}
    \caption{Spearman correlation coefficient of influence scores versus wall-clock time. Hollow markers and solid markers share the same correlation values. Hollow markers only accounts for the time on obtaining pair-wise influence estimates, while solid ones additionally account for the overhead.}
    \label{fig:scalability_validation_2d_spearman}
\end{figure}

\paragraph{Dataset and Model.}
We train a custom GPT-NeoX model on The Penn Treebank dataset~\cite{marcus1993building}. The model consists of decoder layers, two attention heads, and a hidden size of 256, comprising approximately 1.18M MLP parameters and 0.39M MHA parameters.

\paragraph{Baselines.}
We evaluate various TDA methods as baselines. For influence estimation methods relying on iterative iHVP estimation, we include \textit{Conjugate Gradient (CG)} and \textit{LiSSA} following Koh and Liang~\shortcite{koh2017understanding}. To reduce computational costs, we use \textit{CG} as the ground truth instead of methods like training under the proximal Bregman response objective~\cite{bae2022if} or linear datamodeling scores~\cite{park2023trak}.

For IF estimation methods based on estimating diagonal entries of the Hessian, we employ IF with \textit{Arnoldi iteration}~\cite{schioppa2022scaling}.

\textit{TRAK}~\cite{park2023trak}, a retraining-based baseline, uses a set of models trained on random subsets. We also choose \textit{Gradient dot product (GDP)}~\cite{charpiat2019input,grosse2023studying} and \textit{linear Centered Kernel Alignment (CKA)}~\cite{kornblith2019similarity} as gradient-similarity-based baselines. \textit{GDP} simply computes the dot product between query gradients and candidate gradients. \textit{CKA} is specifically designed for measuring representation similarity. Here, we use it on query and training gradients.

\paragraph{Metrics.}
We randomly select ten test samples, each paired with 500 candidates sampled from the training split. The primary metric for influence estimation accuracy is the Spearman correlation coefficient ($\rho$), as ranking quality is our main focus. ``Pair-wise runtime'' refers to the cost of computing the influence of 500 candidates on a query. ``Overhead runtimes'' are specific to each baseline: fitting EK-FAC factors for EK-FAC, calculating dominant eigenpairs for \textit{Arnoldi}, or training and gradient featurization for \textit{TRAK}. Spearman correlations and pair-wise runtimes are averaged over ten trials, while overhead runtime is measured once per baseline.

\begin{table}
{\footnotesize
    \centering
    \renewcommand{\arraystretch}{1.25}
    \setlength{\tabcolsep}{5pt}
    \begin{tabular}{cccc}
    \toprule
    \bf \multirow{2}{*}{\minitab[c]{Method}} & \multicolumn{3}{c}{\bf {Metrics}} \\
    \cline{2-4} & {\bf \makecell{Spearman\\$\rho$ $\uparrow$}} & \bf \makecell{Overhead\\Time $\downarrow$} & {\bf \makecell{Pair-wise\\Time $\downarrow$}} \\ \midrule
    CG& --&0&913.878 s\\
    LiSSA& 0.518&0&1.254 h\\
    \makecell{GDP} & 0.219&0&5.847 s\\
    CKA&0.184&0&6.444 s\\
    \makecell{Arnoldi}& 0.071&3.353 h&235.227 s\\
    TRAK& 0.013&18.633 h&\textbf{0.047 s}\\ \midrule
    EK-FAC& \textbf{0.612}&1.168 h&3.574 s\\
    EK-FAC(MLP)&\underline{0.523}&1.141 h&\underline{2.645 s}\\
    \bottomrule
    \end{tabular}
    \caption{Summary of influence estimation quality and runtime. The \textbf{best} results are highlighted in bold while \underline{second best} results are underlined.}
    \label{tab:scale_validation-transformer}
}
\end{table}

\paragraph{Results.}
Figure \ref{fig:scalability_validation_2d_spearman} visualizes results, with the best results shown in Table \ref{tab:scale_validation-transformer}. Ideally, a method should achieve high-quality estimates within minimal runtime, corresponding to markers closer to the upper left corner of the figure. Key findings include:
\begin{enumerate}
    \item EK-FAC achieves the best trade-off between approximation quality and computation cost, with its marker closest to the upper left corner. Furthermore, it resides on a more optimal Pareto frontier than \textit{CG} and \textit{LiSSA}.
    \item Ablating influence analysis for MHA parameters does not result in substantial degradation in approximation quality compared to analyzing both MHA and MLP parameters. Despite MHA parameters accounting for 25\% of analyzed parameters, they contribute only 14.5\% of the total influence, indicating that MLP parameters have a more significant impact.
    \item The approximation quality of \textit{CG} and \textit{LiSSA} scales log-linearly with computation time, with \textit{CG} offering a superior Pareto frontier.
    \item Gradient similarity-based methods (\textit{GDP} and \textit{CKA}) are the most efficient baselines but yield low-quality influence estimates.
    \item \textit{TRAK}, a representative of retraining-based methods, has the lowest estimation quality and the highest computational cost.
    \item \textit{Arnoldi} yields poor estimates at $\sim$120$\times$ the pair-wise compute cost of EK-FAC.
\end{enumerate}

Observation (2) highlights the potential for further approximations in influence estimation. As noted by Grosse et al.~\shortcite{grosse2023studying}, factual associations are primarily localized within MLP modules~\cite{meng2022locating}, thus MLP modules are also likely to contribute significantly to influence estimation. Our findings support this claim, suggesting that for large-scale models, focusing solely on MLP parameters can significantly reduce computational costs while maintaining useful influence estimates.

\begin{figure}
    \centering
    \begin{subfigure}{1.\linewidth}
        \centering
        \includegraphics[width=1.\linewidth]{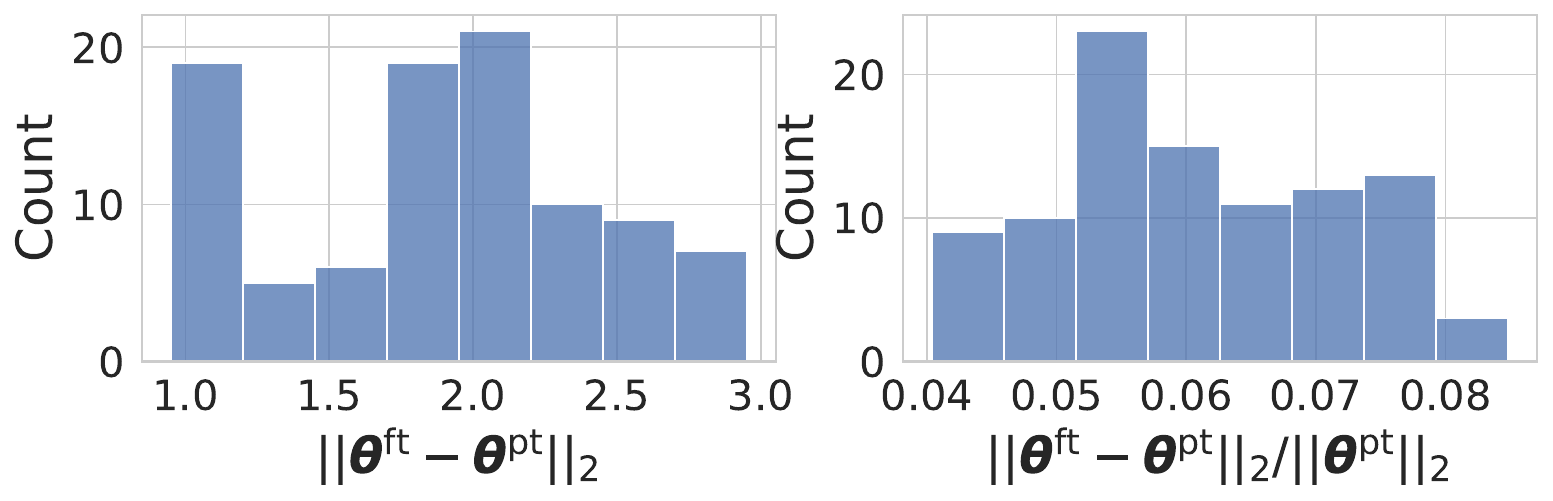}
        \caption{BLOOM-560m v.s. BLOOMZ-560m.}
    \end{subfigure}
    \begin{subfigure}{1.\linewidth}
        \centering
        \includegraphics[width=1.\linewidth]{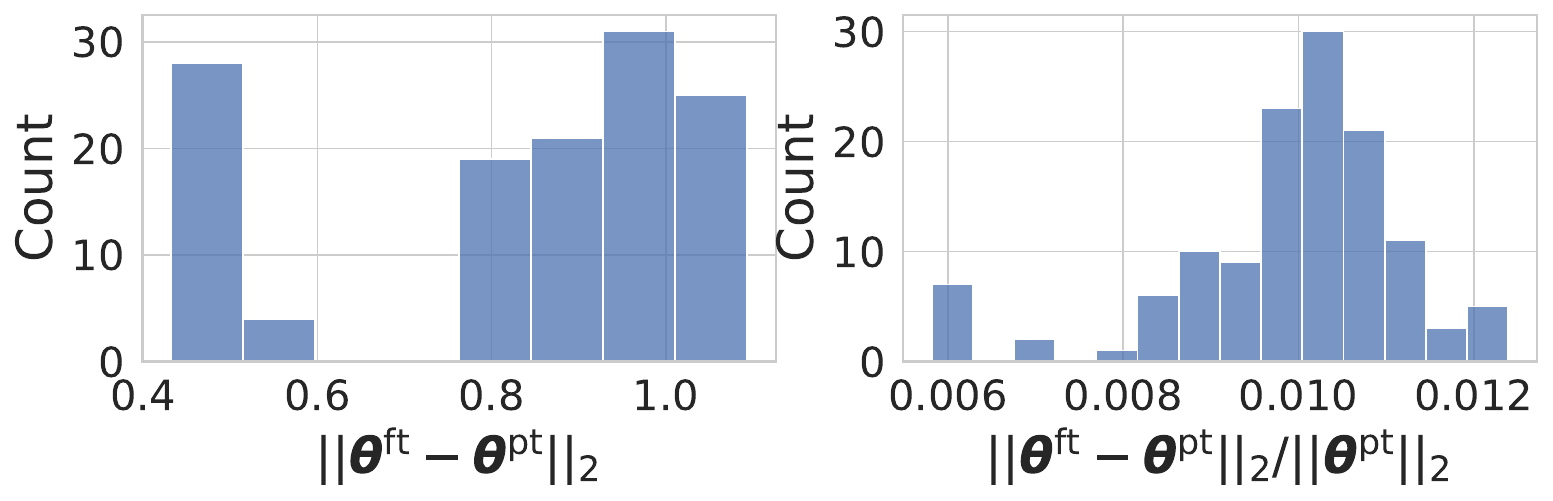}
        \caption{Pythia-2.8b v.s. dolly-v2-3b.}
    \end{subfigure}
    \caption{Distribution of $||\boldsymbol{\theta}^{\text{ft}}-\boldsymbol{\theta}^{\text{pt}}||_2$ and $||\boldsymbol{\theta}^{\text{ft}}-\boldsymbol{\theta}^{\text{pt}}||_2 / ||\boldsymbol{\theta}^{\text{pt}}||_2.$}
    \label{fig:parameter_norm}
\end{figure}

\subsection{Euclidean Proximity in Practice} \label{sec:examine_euclidean_proximity}
As described in Section \ref{sec:multi_stage_influence_function}, the multi-stage IF relies on the assumption that the fine-tuned parameters are geometrically close to its pre-trained predecessor in the parameter space. To validate this assumption, we conduct an experiment analyzing two pairs of models: BLOOM-560m versus BLOOMZ-560m, and Pythia-2.8b versus dolly-v2-3b. We focus on the linear weights of both the MLP and MHA modules, ignoring bias terms, as the bias parameters only constitute a minor portion of the total parameters.

The results, presented in Figure \ref{fig:parameter_norm}, indicate that the distance between fine-tuned and pre-trained weights ($|| \boldsymbol{\theta}^{\text{ft}} - \boldsymbol{\theta}^{\text{pt}} ||_2$) accounts for no more than $8\%$ of the L2 norm of the pre-trained weights, $|| \boldsymbol{\theta}^{\text{pt}} ||_2$. Specifically, for the BLOOM-560m and BLOOMZ-560m pair, $|| \boldsymbol{\theta}^{\text{ft}} - \boldsymbol{\theta}^{\text{pt}} ||_2^2 = 368.7$, while for the Pythia-2.8b and dolly-v2-3b pair, $|| \boldsymbol{\theta}^{\text{ft}} - \boldsymbol{\theta}^{\text{pt}} ||_2^2 = 94.5$.

In practice, the damping term of the GGNs is typically very small; in subsequent experiments, we set $\alpha \leq \lambda_{\text{ft}} = 10^{-4}$. For the two model pairs, the additional Euclidean proximity term introduces an extra fine-tuning loss of $0.0184$ and $0.0047$, respectively. These additional losses are negligible compared to the final training losses of BLOOMZ-560m and dolly-v2-3b, which are $1.6132$ and $0.8209$, respectively.

\subsection{Effectiveness of Multi-Stage Influence Function} \label{sec:multi_stage_infl_effectiveness}
In this experiment, we evaluate the effectiveness of our multi-stage IF in identifying pre-training data that significantly influence predictions of a fine-tuned model. To facilitate this evaluation, we establish a benchmark with ground-truth labels for attributing test-set instances to corresponding samples in an attribution set. To enable comparison with the single-stage IF, the chosen task does not require modifying the unembedding layer.

\begin{figure}
    \centering
    \includegraphics[width=.9\linewidth]{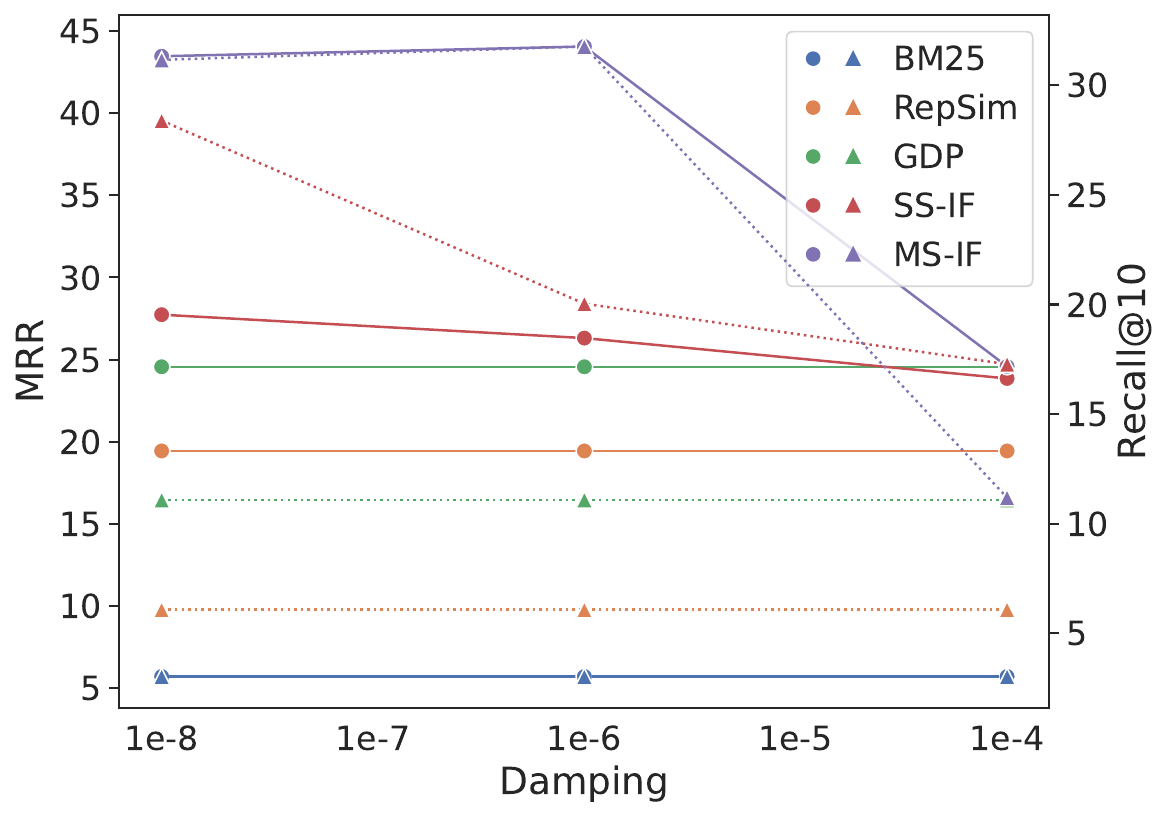}
    \caption{Fact-tracing results (in percentage). Round markers and solid lines denote MRR results, while triangular markers and dotted lines denote Recall@10 results.}
    \label{fig:fact_tracing}
\end{figure}

\paragraph{Benchmark Setup.}
We construct a fact-tracing benchmark based on that of Aky\"urek et al.~\shortcite{akyurek2022towards}, consisting of an attribution set and a test set. Each test set instance corresponds to a fact and is associated with several sentences in the attribution set. The evaluated methods are tasked with correctly retrieving the knowledge sources for each fact.

The attribution set is derived from the T-REx dataset~\cite{elsahar2018t}, which aligns knowledge base triples with DBpedia abstracts. Instead of directly using the original dataset, we create a sentence-level subset of T-REx that includes all knowledge base triples represented in the LAMA dataset~\cite{petroni2019language}.

The test set is derived from the T-REx split of LAMA. Aky\"urek et al.~\shortcite{akyurek2022towards} use the test set as cloze-style language modeling samples, as their target models are mT5-based~\cite{xue2020mt5} models, which are trained under the masked language modeling “span-corruption” objective. But our target model is an instruction-following autoregressive LLM, so we manually convert the original text completion templates into equivalent question answering formats. For example, a knowledge base triple (X, born\_in, Y) originally uses the template “X was born in,” which is a text-completion task, while our version reformulates it as “Where was X born?” to align with a question-answering task, where the model is required to predict the object Y.

\paragraph{Metrics.}
We evaluate fact retrieval performance using standard information retrieval metrics, including Mean Reciprocal Rank (MRR) and Recall@10. The MRR is defined as $\frac{1}{Q}\sum_{q \in Q}\frac{1}{\text{rank}_q}$, where $Q$ is the test set and $\text{rank}_q$ denotes the rank of the first correct knowledge source for query $q$. Results are averaged over three trials, with each trial sampling 200 test instances.

\paragraph{Baselines.}
We compare the proposed multi-stage IF (\textit{MS-IF}) against several baseline methods.
\textit{BM25}~\cite{robertson1995okapi}, a model-agnostic baseline, retrieves relevant samples based on word-level overlap.
Additionally, we include two similarity-based methods: Representation Similarity (\textit{RepSim})~\cite{caruana1999case}, which computes cosine similarity between the hidden states of pre-trained and fine-tuned models, and \textit{GDP}, a simplified multi-stage IF where the Hessian matrices are identity matrices.

The single-stage IF (\textit{SS-IF}) is also evaluated, using only the fine-tuned model. Both \textit{MS-IF} and \textit{SS-IF} adopt the same measurement $m(x, y; \boldsymbol{\theta}) = -\log p_{\boldsymbol{\theta}}(y|x)$, where $x$ and $y$ represent the question and answer tokens, respectively. Both use EK-FAC approximation and analyze both the MLP and MHA modules. Furthermore, as both IFs rely on damping terms, we report their performance under different damping values, ranging from $10^{-8}$ to $10^{-4}$.

\paragraph{Models.}
We utilize BLOOM-560m~\cite{workshop2022bloom} and BLOOMZ-560m~\cite{muennighoff2022crosslingual}, the latter of which has undergone multitask fine-tuning. We choose these models as we are able to verify that their pre-training data encompass most of the knowledge of the fact-tracing benchmark. Specifically, as each knowledge instance is represented as (subject, relation, object) triples, we confirm that 96.81\% of the objects are included by the pre-training data of BLOOM(Z)-560m.

\paragraph{Results.}
As is shown in Figure \ref{fig:fact_tracing}, the multi-stage IF outperforms all baseline methods in terms of both MRR and Recall under smaller damping terms ($10^{-6}$ and $10^{-8}$), demonstrating its superior ability to assign higher influence scores to ground-truth knowledge sources. However, the performance gaps between \textit{MS-IF}, \textit{SS-IF}, and \textit{GDP} are relatively small under a large damping term ($10^{-4}$). This indicates that the choice of the damping term is essential to the effectiveness of IFs, and that both \textit{SS-IF} and \textit{MS-IF} degenerate to \textit{GDP} when the damping terms are large.
Moreover, even the best-performing method yields results that are far from perfect retrieval. This discrepancy may stem from approximation errors, the design of our multi-stage IF, or limitations in the language models’ suitability for knowledge retrieval tasks.

\subsection{Case Study}\label{sec:qualitative_studies}
In addition to the quantitative experiments above, we qualitatively demonstrate the interpretive power of the proposed multi-stage IF through a case study. The analysis is conducted on dolly-v2-3b for the task of factual knowledge attribution.
The motivation for this case study stems from possible user concerns regarding whether an LLM-based interactive system's responses are grounded in reliable knowledge sources or are mere hallucinations. To address this, it is reasonable to attribute model-generated outputs to the pre-training data. By inspecting whether the top-influential pre-training texts contain relevant and accurate information, we are able to assess whether the model's responses are properly grounded.

For the factual knowledge attribution task, the model is prompted with a question, and a response is generated using greedy decoding. The article with the highest multi-stage influence is identified from the Wikipedia subset of the pre-training corpus using the pipeline described in Section \ref{sec:method_body}. Table \ref{tab:fact_tracing_example_short} presents the results, showing an excerpt of the retrieved article. While the response itself is incorrect, the retrieved article is highly relevant to the queried topic.

\begin{table}
{\footnotesize
    \centering
    \renewcommand{\arraystretch}{1.25}
    \setlength{\tabcolsep}{5pt}
    \begin{tabular}{rl}
    \toprule
    \bf{Query} & \makecell[l]{Q: Where did fortune cookies originate?\\A: The fortune cookie originated in China.}\\ \midrule
    \bf \makecell{Retrieved\\Document} & \makecell[l]{Fortune cookies are often served as a \\dessert in Chinese restaurants in the United \\States and other Western countries, but are\\ not a tradition in China. [...]As far back as\\ the 19th century, a cookie very similar in \\appearance to the modern fortune cookie \\was made in Kyoto, Japan [...]} \\
    \bottomrule
    \end{tabular}
    \caption{Example of the most influential example from the pre-training dataset for a fact-related query.}
    \label{tab:fact_tracing_example_short}
}
\end{table}

\section{Conclusions and Limitations}
This paper introduces a generalization of the IF to enable the attribution of predictions made by a fine-tuned model to its pre-training data. To enhance the scalability of influence computation, we employ EK-FAC parameterization and a nearest-neighbor-based candidate selection strategy. Experimental results confirm the effectiveness and efficiency of the proposed multi-stage IF, demonstrating its applicability to LLMs with three billion parameters.

While our work enhances the scalability and generality of influence functions, several limitations remain.
First, we only analyze backbone MLP and MHA components, excluding contributions from other components such as embeddings, unembeddings, and layer normalization. Extending influence analysis to these components may improve the quality of influence estimates. However, the potential benefit is likely marginal, as they constitute a small proportion of the overall parameters and are not considered to encode knowledge.

Second, our analyses are limited to decoder-only transformer architectures. Extending scalable influence analysis to other architectures, such as encoder-decoder models or diffusion models, could unlock valuable new applications.

Finally, \textsc{Source} is proposed by Bae et al.~\shortcite{bae2024training} as an effective TDA approach, leveraging approximate unrolled differentiation and inherently suited for multi-stage scenario. Their results demonstrate the superior performance of \textsc{Source} compared to influence functions on models like BERT~\cite{devlin2018bert} and GPT-2~\cite{radford2019language}. Unfortunately, due to the unavailability of its implementation, we were unable to directly compare our multi-stage influence function with \textsc{Source} in this work.

\section*{Acknowledgments}
This work was partly supported by the National Key Research and Development Program of China under No. 2024YFB3900105, NSFC under No. 62402418, Zhejiang Province's 2025 “Leading Goose + X” Science and Technology Plan under grant No.2025C02034, the Key R\&D Program of Ningbo under No. 2024Z115, and the Open Project of Key Laboratory of General Quality Technology and Application of Intelligent Manufacturing Equipment, Ministry of Industry and Information Technology (HK202403532).

\bibliographystyle{named}
\bibliography{ijcai25}

\appendix

\section{Proof of EK-FAC}
\begin{proof}
Following the approach of Grosse et al.~\shortcite{grosse2023studying}, we approximate the GGN matrix ($\mathbf{G}$) using EK-FAC parameterization. We organize our proof in three steps: (1) approximate GGN using block-wise Kronecker-factored approximation as in the work of Martens and Grosse~\shortcite{martens2020optimizing}; (2) additionally approximate GGN as block-diagonal for efficient inversion; (3) EK-FAC to improve approximation quality and to enable efficient inverse Gauss-Newton Hessian-vector product (iGNHVP), $(\mathbf{G}+\lambda \mathbf{I})^{-1}\mathbf{v}$. The first two steps trade approximation quality for efficiency while the third step partially supplements approximation quality.

First we present the derivation for the GGN matrix, which is commonly used to approximate the Hessian~\cite{eschenhagen2024kronecker}:
\begin{equation}
\begin{aligned}
    \mathbf{G} = & \underset{x \sim Q_x}{\mathbb{E}} \big[ \nabla_{\boldsymbol{\theta}}\log p(y|f_{\boldsymbol{\theta}}(x),\boldsymbol{\theta}) \\
    & \nabla_{\boldsymbol{\theta}}\log p(y|f_{\boldsymbol{\theta}}(x),\boldsymbol{\theta})^\top \big],
\end{aligned}
\end{equation}
where $Q_x$ is the training data distribution, $\boldsymbol{\theta}$ is model parameters and $y$ is model outputs. In the case of cross-entropy loss, GGN is equivalent to the Fisher Information matrix. As $Q_x$ is infinite and inaccessible, the GGN is usually approximated with the empirical mean over the distribution of the training set $\hat{Q}_x$:
\begin{equation}
\begin{aligned}
    \mathbf{G} = &\underset{\begin{subarray}{c}
    (x_n,y_n)\sim \hat{Q}_x\\
    y\sim p(y|f_{\boldsymbol{\theta}}(x_n))
    \end{subarray}}{\mathbb{E}} \Bigl[ \mathcal{D}\boldsymbol{\theta} \mathcal{D}\boldsymbol{\theta}^\top \Bigr],
\end{aligned}
\end{equation}
where $\mathcal{D}\boldsymbol{\theta} = \nabla_{\boldsymbol{\theta}}\log p(y|f_{\boldsymbol{\theta}}(x_n),\boldsymbol{\theta})$.
Note that instead of using labels from $\hat{Q}_x$, the labels are sampled from the model's output distribution.

\paragraph{1. Block-wise Kronecker-factored approximation.}
Suppose $\boldsymbol{\theta}_1, ... \boldsymbol{\theta}_L$ correspond to the parameters of a model's sub-networks. For example, they might be the parameters of an MLP with $L$ layers. Now we focus on a single linear sub-network, $\boldsymbol{\theta}_l$, whose inputs and outputs are $\mathbf{a}_{l-1}$ and $\mathbf{s}_l$. For simplicity, we assume the linear sub-networks have no bias term. However, the additional bias could be absorbed as an additional column of the weight matrix, plus appending ``1'' at the end of $\mathbf{a}_{l-1}$. The pseudo-gradients $\mathcal{D}\boldsymbol{\theta}_l$ can be expressed as a Kronecker product:
\begin{equation}
\begin{aligned}
    \mathcal{D}\boldsymbol{\theta}_l = \mathbf{a}_{l-1} \otimes \mathcal{D}\mathbf{s}_l.
\end{aligned}
\end{equation}

The GGN for $\boldsymbol{\theta}_l$ can then be expressed as:
\begin{equation}
\begin{aligned}
    \mathbf{G}(\boldsymbol{\theta}_l) &= \mathbb{E}\left[ (\mathbf{a}_{l-1}\otimes\mathcal{D}\mathbf{s}_l) (\mathbf{a}_{l-1}\otimes\mathcal{D}\mathbf{s}_l)^\top \right]\\
    &= \mathbb{E}\left[ (\mathbf{a}_{l-1}\mathbf{a}_{l-1}^\top) \otimes (\mathcal{D}\mathbf{s}_l \mathcal{D}\mathbf{s}_l^\top) \right].
\end{aligned}
\end{equation}

Naturally we can derive a block-wise formulation for the GGN:
\begin{equation}
\begin{aligned}
    \mathbf{G}_{i,j} &=  \mathbb{E}\left[ \mathcal{D}\boldsymbol{\theta}_i \mathcal{D}\boldsymbol{\theta}_j^\top \right] \\
    &= \mathbb{E}\left[ (\mathbf{a}_{i-1}\mathbf{a}_{j-1}^\top) \otimes (\mathcal{D}\mathbf{s}_i \mathcal{D}\mathbf{s}_j^\top) \right],
\end{aligned}
\end{equation}
where $i,j=1,...,L$ and $G_{i,j}$ corresponds to the $i$th row, $j$th column block of the GGN.

Under the assumption that inputs $\mathbf{a}_{l-1}$ and output pseudo-gradients $\mathcal{D}\mathbf{s}_l$ are independent, we can obtain the following approximation:
\begin{equation}\label{eq:kfac}
\begin{aligned}
    \mathbf{G}_{i,j} &\approx \mathbb{E}\left[ (\mathbf{a}_{i-1}\mathbf{a}_{j-1}^\top)\right] \otimes \mathbb{E}\left[(\mathcal{D}\mathbf{s}_i \mathcal{D}\mathbf{s}_j^\top) \right]\\
    &= \mathbf{A}_{i-1,j-1} \otimes \mathbf{S}_{i,j},
\end{aligned}
\end{equation}
where $\mathbf{A}_{i-1,j-1}$ and $\mathbf{S}_{i,j}$ are covariance matrices. Equation (\ref{eq:kfac}) results in a block-wise Kronecker-Factored (K-FAC) approximation of the GGN.

\paragraph{2. Approximating GGN as block-diagonal.} However, eventually we need to compute $(\mathbf{G}+\lambda \mathbf{I})^{-1}\mathbf{v}$, which requires inverting a damped GGN. The block-wise K-FAC approximation does not allow for efficient inversion of the GGN. By only preserving the block-diagonal entries of $\mathbf{G}$, we could trivially invert $\mathbf{G}+\lambda \mathbf{I}$ by inverting each of the diagonal blocks. Therefore following Martens~\shortcite{martens2020new} and Grosse et al.~\shortcite{grosse2023studying}, we adopt an additional block-diagonal approximation of GGN:
\begin{equation}
\begin{aligned}
    \mathbf{G}(\boldsymbol{\theta}) = \text{diag}(\mathbf{G}(\boldsymbol{\theta}_1), \mathbf{G}(\boldsymbol{\theta}_2), ..., \mathbf{G}(\boldsymbol{\theta}_L)),
\end{aligned}
\end{equation}
where $\mathbf{G}(\boldsymbol{\theta}_l)=\mathbf{G}_{l,l}$.

In the case of transformer models, we take a finer granularity than layer-wise approximation. In our approximation, each ``block'' corresponds to the parameters of a linear projection matrix, which corresponds to MLP weights and linear projection weights of MHA modules.

\paragraph{3. EK-FAC: Improve K-FAC by tracking diagonal variance.} Because of the independence hypothesis induced by the K-FAC approximation, the approximation can be further improved. Next, we introduce EK-FAC, which supplements the correlation between $\mathbf{A}_{l-1}$ and $\mathbf{S}_l$. Applying eigen-decomposition on these two and omitting the block-specific suffix $l$, we obtain $\mathbf{Q_A}$, $\mathbf{\Lambda_A}$, $\mathbf{Q_S}$ and $\mathbf{\Lambda_S}$, which are eigenvectors and eigenvalues for $\mathbf{A}$ and $\mathbf{S}$:
\begin{equation}
\begin{aligned}
    \mathbf{A}\otimes\mathbf{S} &= (\mathbf{Q_A}\mathbf{\Lambda_A}\mathbf{Q_A}^\top) \otimes (\mathbf{Q_S}\mathbf{\Lambda_S}\mathbf{Q_S}^\top) \\
    &= (\mathbf{Q_A}\otimes\mathbf{Q_S}) (\mathbf{\Lambda_A}\otimes\mathbf{\Lambda_S}) (\mathbf{Q_A}\otimes\mathbf{Q_S})^\top.
\end{aligned}
\end{equation}

The eigenvalues $\mathbf{\Lambda_A}$ and $\mathbf{\Lambda_S}$ are discarded. Instead, we fit another diagonal matrix to replace $\mathbf{\Lambda_A}\otimes\mathbf{\Lambda_S}$ by capturing the covariance of gradients projected onto the eigenvectors:
\begin{equation}
\begin{aligned}
    \mathbf{\Lambda}_{ii} &= \mathbb{E}\left[\left( \left(\mathbf{Q_A}\otimes\mathbf{Q_S}\right)^\top \mathcal{D}\boldsymbol{\theta} \right)_i^2\right] \\
    &= \mathbb{E}\left[ \left( \text{vec}\left( \mathbf{Q_S}^\top \mathcal{D}\mathbf{W} \mathbf{Q_A} \right) \right)_i^2 \right],
\end{aligned}
\end{equation}
where $\mathbf{W}$ is vector-form $\boldsymbol{\theta}$ reshaped into matrix form and $\mathbf{\Lambda}_{ii}$ is the $i$-th diagonal element of the new diagonal matrix $\mathbf{\Lambda}$. As we need a damped version of iGNHVP to ensure positive semi-definiteness, a damping term $\lambda$ is added to the diagonal matrix:
\begin{equation}
\begin{aligned}
    \mathbf{\Lambda}_{\lambda} = \mathbf{\Lambda} + \lambda \mathbf{I}.
\end{aligned}
\end{equation}

Now we are ready to derive the iGNHVP:
\begin{equation}
\begin{alignedat}{2}
    \left(\mathbf{G}(\boldsymbol{\theta}_l)+\lambda \mathbf{I}\right)^{-1}\mathbf{v}_l & \approx && (\mathbf{Q_A}\otimes\mathbf{Q_S}) \mathbf{\Lambda}_{\lambda}^{-1} \\
    & && (\mathbf{Q_A}\otimes\mathbf{Q_S})^\top \mathbf{v}_l \\
    & = &&\text{vec} \Bigg(\mathbf{Q_S}\bigg[ \left(\mathbf{Q_S}^\top \bar{\mathbf{V}_l} \mathbf{Q_A}\right) \oslash \\
    & &&\text{unvec}\left(\text{diag}^{-1} \left(\mathbf{\Lambda}_{\lambda}\right)\right)\bigg]\mathbf{Q_A}^\top\Bigg),
\end{alignedat}
\end{equation}
where $\oslash$ denotes element-wise division, $\text{diag}(\cdot)$ denotes filling a vector into diagonal entries of a diagonal matrix, $\text{diag}^{-1}(\cdot)$ denotes extracting the diagonal entries of a matrix to a vector and $\text{unvec}(\cdot)$ denotes converting a vector into two-dimensional matrix form. $\mathbf{v}_l$ is the slice of $\mathbf{v}$ corresponding to the parameter $\boldsymbol{\theta}_l$.

The resultant $(\mathbf{G}(\boldsymbol{\theta})+\lambda \mathbf{I})^{-1}\mathbf{v}$ is obtained by concatenating $(\mathbf{G}(\boldsymbol{\theta}_l)+\lambda \mathbf{I})^{-1}\mathbf{v}_l$ for $l=1,...,L$.
\end{proof}

\subsection{Computation and Memory Cost}
The following analysis is limited to a single decoder layer, which could be trivially extended to the model backbone composed of identical decoder layers.

Denoting the input and output dimensions of a weight matrix by $d$ and $p$, the computation cost of preparing EK-FAC factors consists of two major components: (1) $2\times$ forward and backward passes over samples drawn from the training distribution, one for fitting covariance factors and another for fitting the diagonal, resulting in an asymptotic computational cost of $\mathcal{O}(dpN)$, where $N$ is the number of mini-batches; (2) eigen-decomposition of covariance matrices. Fortunately, these operations are one-time overhead, with the cost amortized across future influence analyses. For influence score derivation, the computational cost of matrix multiplications when computing an iHVP is $\mathcal{O}(d^2p+dp^2)$, followed by a dot product between iGNHVP and candidate gradients, which has a cost of $\mathcal{O}(dp)$.

The memory and storage overhead corresponds to eigenvectors $\mathbf{Q}_{\mathbf{A}}$, $\mathbf{Q}_{\mathbf{S}}$ and diagonal entries of $\mathbf{\Lambda}$. The extra spatial cost is $d^2 + p^2 + dp$.

Take GPT-NeoX architecture as an example, for query/key/value weights and linear projection weights of MHA modules, $d=p$. Their additional spatial cost is approximately 3 times that of the original weights. For weight matrices of MLP modules, either $d=4p$ or $d=\frac{1}{4} p$. Therefore, the extra spatial cost is approximately 5.25 times that of the original weights.

\section{Proof of Multi-Stage Influence Function under Full-Parameter Fine-Tuning} \label{sec:multi_stage_influence_function_proof}
\begin{proof}
In the context of full-parameter fine-tuning, we first formulate the standalone training objectives for the pre-training stage and fine-tuning stage, respectively. During pre-training, all the parameters are initialized randomly and updated to minimize empirical loss on a general-purpose pre-training corpus with $N$ samples:
\begin{equation}\label{eq:pretrain_obj}
\begin{aligned}
    \boldsymbol{\theta}^{\text{pt}} = \argmin_{\boldsymbol{\theta}} \mathcal{L}_{\text{pt}}(\boldsymbol{\theta}) = \argmin_{\boldsymbol{\theta}} \frac{1}{N}\sum_{i=1}^N \ell_{\text{pt}}(z_i, \boldsymbol{\theta}),
\end{aligned}
\end{equation}
where $\ell_{\text{pt}}(\cdot)$ is the pre-training loss, $\mathcal{L}_{\text{pt}}(\cdot)$ is the average pre-training loss and $\boldsymbol{\theta}^{\text{pt}}$ is optimal parameters after pre-training.

Following pre-training, the pre-trained model is fine-tuned on a task-specific dataset with $M$ samples:
\begin{equation}\label{eq:finetune_obj}
\begin{aligned}
    \boldsymbol{\theta}^{\text{ft}} = \argmin_{\boldsymbol{\theta}} \mathcal{L}_{\text{ft}}(\boldsymbol{\theta}) = \argmin_{\boldsymbol{\theta}} \frac{1}{M}\sum_{i=1}^M \ell_{\text{ft}}(x_i, \boldsymbol{\theta}),
\end{aligned}
\end{equation}
where $\ell_{\text{ft}}(\cdot)$ is the fine-tuning loss, $\mathcal{L}_{\text{ft}}(\cdot)$ is the average fine-tuning loss and $\boldsymbol{\theta}^{\text{ft}}$ is optimal parameters after fine-tuning.

Suppose we use the original influence function to study the influence of pre-training sample $z$ on a fine-tuning instance $x$. Then ideally we wish to compute the following formula using only the fine-tuned model, using measurement $m(\cdot)$:
\begin{equation} \label{eq:influence_finetune_only}
\begin{aligned}
    \mathcal{I}_{m}(z, x) = \nabla_{\boldsymbol{\theta}} m(x, \boldsymbol{\theta}^{\text{ft}})^\top \mathbf{H}_{\boldsymbol{\theta}^{\text{ft}}}^{-1} \nabla_{\boldsymbol{\theta}} \ell_{\text{pt}}(z, \boldsymbol{\theta}^{\text{ft}}).
\end{aligned}
\end{equation}

The challenge arises for the original influence function when $\ell_{\text{pt}}$ and $\ell_{\text{ft}}$ are different, which makes the fine-tuned model unable to yield the pre-training gradient $\nabla_{\boldsymbol{\theta}} \ell_{\text{pt}}(z, \boldsymbol{\theta}^{\text{ft}})$. For example, support we train a language model using autoregressive pre-training loss $\ell_{\text{pt}}(z, \boldsymbol{\theta}) = -\sum_{i=1}^{n}\log p_{\boldsymbol{\theta}}(z_i|z_{j<i}, \boldsymbol{\theta})$. Then we fine-tuned the model for a text classification task by replacing its language modeling head with a classification head, under fine-tuning loss to maximize the predictive probability of label $y$: $\ell_{\text{ft}}(x, y, \boldsymbol{\theta}) = -\log p_{\boldsymbol{\theta}} (y|x, \boldsymbol{\theta})$. As the final layer of fine-tuned model is dedicated to the downstream classification task rather than the language modeling task, we can no longer compute gradients under pre-training loss using the fine-tuned model. Therefore when the final prediction layer is replaced, Equation (\ref{eq:influence_finetune_only}) is a false formulation.

In order to make the pre-training gradients accessible for the fine-tuned model, we need to establish a connection between these two training stages. Same as Chen et al.~\shortcite{chen2020multi}, we take advantage of the proximity in parameter space between the pre-trained model and the fine-tuned model to reformulate the joint training objective as follows:
\begin{equation}
\begin{aligned}
    \boldsymbol{\theta}^{\text{ft}} = \argmin_{\boldsymbol{\theta}} \mathcal{L}_{\text{ft}}(\boldsymbol{\theta}) + \alpha ||\boldsymbol{\theta}-\boldsymbol{\theta}^{\text{pt}}||_2^2,
\end{aligned}
\end{equation}
where $\alpha$ is a constant satisfying $0 < \alpha \leq 1$. This formulation explicitly takes account of the assumption that fine-tuned model parameters are located in the adjacency of pre-trained model parameters. Furthermore, this joint formulation can be viewed as enforcing L2 regularization during fine-tuning.

Suppose we are analyzing the influence of a pre-training sample $z$ on a fine-tuning test sample $x_t$. Following formulations from Koh and Liang~\shortcite{koh2017understanding}, the pre-training instance $z$ is to be perturbed with strength $\epsilon$ ($0 < |\epsilon| \ll 1$), resulting in perturbations in the pre-training objective:
\begin{equation}\label{eq:pretrain_obj_perturb}
\begin{aligned}
    \boldsymbol{\theta}^{\text{pt}}({\epsilon}) &= \argmin_{\boldsymbol{\theta}} \mathcal{L}_{\text{pt}}(\boldsymbol{\theta}) + \epsilon \ell_{\text{pt}}(z, \boldsymbol{\theta}),
\end{aligned}
\end{equation}
where $\boldsymbol{\theta}_{\text{pt}}({\epsilon})$ is the optimal solution under the pre-training objective from a perturbation of strength $\epsilon$ on pre-training sample $z$. Letting $\epsilon=-1/N$ amounts to exactly removing $z$ from the pre-training dataset.

\begin{equation}\label{eq:finetune_obj_perturb}
\begin{aligned}
    \boldsymbol{\theta}^{\text{ft}}({\epsilon}) &= \argmin_{\boldsymbol{\theta}} \mathcal{L}_{\text{ft}}(\boldsymbol{\theta}) + \frac{\alpha}{2} ||\boldsymbol{\theta}-\boldsymbol{\theta}^{\text{pt}}({\epsilon})||^2,
\end{aligned}
\end{equation}
where $\boldsymbol{\theta}^{\text{ft}}(\epsilon)$ is the optimal solution under the reformulated fine-tuning objective from $\epsilon$ perturbation on the pre-training sample $z$.

Since $\boldsymbol{\theta}^{\text{pt}}({\epsilon})$ and $\boldsymbol{\theta}^{\text{ft}}({\epsilon})$ are optimal solutions, Equation (\ref{eq:finetune_obj_perturb}) gives rise to the following optimal condition:
\begin{equation}\label{eq:finetune_optimal}
\begin{aligned}
    \nabla_{\boldsymbol{\theta}} \mathcal{L}_{\text{ft}}(\boldsymbol{\theta}^{\text{ft}}({\epsilon})) + \frac{\alpha}{2} \nabla_{\boldsymbol{\theta}} || \boldsymbol{\theta}^{\text{ft}}({\epsilon}) - \boldsymbol{\theta}^{\text{pt}}({\epsilon}) ||^2 = 0.
\end{aligned}
\end{equation}

Similarly, Equation (\ref{eq:pretrain_obj_perturb}) derives the following optimal condition:
\begin{equation}\label{eq:pretrain_optimal}
\begin{aligned}
    \nabla_{\boldsymbol{\theta}} \mathcal{L}_{\text{pt}}(\boldsymbol{\theta}^{\text{pt}}({\epsilon})) + \epsilon \nabla_{\boldsymbol{\theta}} \ell_{\text{pt}}(z, \boldsymbol{\theta}^{\text{pt}}({\epsilon})) = 0.
\end{aligned}
\end{equation}

Applying first-order Taylor expansion on Equation (\ref{eq:pretrain_optimal}) results in:
\begin{equation}
\begin{aligned}
    0 = &\nabla_{\boldsymbol{\theta}} \mathcal{L}_{\text{pt}}(\boldsymbol{\theta}^{\text{pt}}({\epsilon})) + \nabla_{\boldsymbol{\theta}}^2 \mathcal{L}_{\text{pt}}(\boldsymbol{\theta}^{\text{pt}}({\epsilon})) \Delta \boldsymbol{\theta}^{\text{pt}}({\epsilon}) \\
    &+ \epsilon \nabla_{\boldsymbol{\theta}} \ell_{\text{pt}}(z, \boldsymbol{\theta}^{\text{pt}}({\epsilon})) \\
    &+ \epsilon \nabla_{\boldsymbol{\theta}}^2 \ell_{\text{pt}}(z, \boldsymbol{\theta}^{\text{pt}}({\epsilon})) \Delta \boldsymbol{\theta}^{\text{pt}}({\epsilon}),
\end{aligned}
\end{equation}
where $\Delta \boldsymbol{\theta}^{\text{pt}}({\epsilon}) = \boldsymbol{\theta}^{\text{pt}}({\epsilon}) - \boldsymbol{\theta}^{\text{pt}}$.

As $\boldsymbol{\theta}^{\text{pt}}$ is an optimal solution for the unperturbed pre-training objective according to Equation (\ref{eq:pretrain_obj}), $\nabla_{\boldsymbol{\theta}} \mathcal{L}_{\text{pt}}(\boldsymbol{\theta}^{\text{pt}}({\epsilon}))=0$. Additionally, $\epsilon \rightarrow 0$. It is natural to obtain:
\begin{equation}\label{eq:pretrain_delta}
\begin{aligned}
    \Delta \boldsymbol{\theta}^{\text{pt}}({\epsilon}) = -\epsilon (\nabla_{\boldsymbol{\theta}}^2 \mathcal{L}_{\text{pt}}(\boldsymbol{\theta}^{\text{pt}}({\epsilon})))^{-1} \nabla_{\boldsymbol{\theta}} \ell_{\text{pt}}(z, \boldsymbol{\theta}^{\text{pt}}({\epsilon})).
\end{aligned}
\end{equation}

Again, applying first-order Taylor expansion on Equation (\ref{eq:finetune_optimal}) results in:
\begin{equation}
\begin{aligned}
    0 = &\nabla_{\boldsymbol{\theta}} \mathcal{L}_{\text{ft}}(\boldsymbol{\theta}^{\text{ft}}({\epsilon})) + \nabla_{\boldsymbol{\theta}}^2 \mathcal{L}_{\text{ft}}(\boldsymbol{\theta}^{\text{ft}}) \Delta \boldsymbol{\theta}^{\text{ft}}({\epsilon}) \\
    &+ \frac{\alpha}{2} \nabla_{\boldsymbol{\theta}} || \boldsymbol{\theta}^{\text{ft}} - \boldsymbol{\theta}^{\text{pt}}({\epsilon}) ||^2 \\
    &+ \frac{\alpha}{2} \nabla_{\boldsymbol{\theta}}^2 || \boldsymbol{\theta}^{\text{ft}} - \boldsymbol{\theta}^{\text{pt}}({\epsilon}) ||^2 \Delta \boldsymbol{\theta}^{\text{pt}}({\epsilon}),
\end{aligned}
\end{equation}
where $\Delta \boldsymbol{\theta}^{\text{ft}}({\epsilon}) = \boldsymbol{\theta}^{\text{ft}}({\epsilon}) - \boldsymbol{\theta}^{\text{ft}}$.

As $\boldsymbol{\theta}^{\text{ft}}$ is an optimal solution on the unperturbed fine-tuning objective as per Equation (\ref{eq:finetune_obj}), $\nabla_{\boldsymbol{\theta}} \mathcal{L}_{\text{ft}}(\boldsymbol{\theta}^{\text{ft}}) = 0$. Meanwhile, as $\boldsymbol{\theta}^{\text{pt}}({\epsilon})$ is a constant at fine-tuning stage, $\nabla_{\boldsymbol{\theta}}^2 || \boldsymbol{\theta}^{\text{ft}} -  \boldsymbol{\theta}^{\text{pt}}({\epsilon})||^2 = 2\mathbf{I}$. Then:
\begin{equation}\label{eq:finetune_delta_before}
\begin{aligned}
    \Delta \boldsymbol{\theta}^{\text{ft}}({\epsilon}) &= &&-(\nabla_{\boldsymbol{\theta}}^2 \mathcal{L}_{\text{ft}}(\boldsymbol{\theta}^{\text{ft}}) + 2\alpha \mathbf{I})^{-1} \\
    & && \Bigl[ \nabla_{\boldsymbol{\theta}}\mathcal{L}_{\text{ft}}(\boldsymbol{\theta}^{\text{ft}}) + 2\alpha || \boldsymbol{\theta}^{\text{ft}} - \boldsymbol{\theta}^{\text{pt}}({\epsilon}) || \Bigr] \\
    &= &&-2\alpha(\nabla_{\boldsymbol{\theta}}^2 \mathcal{L}_{\text{ft}}(\boldsymbol{\theta}^{\text{ft}}) + 2\alpha \mathbf{I})^{-1} \\
    & && || \boldsymbol{\theta}^{\text{ft}} - \boldsymbol{\theta}^{\text{pt}}({\epsilon}) ||.
\end{aligned}
\end{equation}

Take the result from Equation (\ref{eq:pretrain_delta}) into Equation (\ref{eq:finetune_delta_before}) with the help of  $\boldsymbol{\theta}^{\text{pt}}({\epsilon}) = \boldsymbol{\theta}^{\text{pt}} + \Delta \boldsymbol{\theta}^{\text{pt}}({\epsilon})$, and compute its derivative respect to $\epsilon$ at $\epsilon \rightarrow 0$. Ignoring factors without the term $\epsilon$, we can get
\begin{equation}\label{eq:finetune_delta_after}
\begin{aligned}
    \frac{\partial \Delta \boldsymbol{\theta}^{\text{ft}}({\epsilon})}{\partial \epsilon} \bigg|_{\epsilon =0} &= && \frac{\partial}{\partial \epsilon} \bigg[ \alpha \left(\nabla_{\boldsymbol{\theta}}^2 \mathcal{L}_{\text{ft}}\left(\boldsymbol{\theta}^{\text{ft}}\right) + \alpha\mathbf{I}\right)^{-1} \\
    & &&\epsilon (\nabla_{\boldsymbol{\theta}}^2 \mathcal{L}_{\text{pt}}(\boldsymbol{\theta}^{\text{pt}}))^{-1} \nabla_{\boldsymbol{\theta}} \ell_{\text{pt}}(\boldsymbol{\theta}^{\text{pt}})\bigg] \\
    &= &&\alpha \left(\nabla_{\boldsymbol{\theta}}^2 \mathcal{L}_{\text{ft}}\left(\boldsymbol{\theta}^{\text{ft}}\right) + \alpha\mathbf{I}\right)^{-1}\\
    & &&\left(\nabla_{\boldsymbol{\theta}}^2\mathcal{L}_{\text{pt}}\left(\boldsymbol{\theta}^{\text{pt}}\right)\right)^{-1} \nabla_{\boldsymbol{\theta}} \ell_{\text{pt}}(\boldsymbol{\theta}^{\text{pt}}).
\end{aligned}
\end{equation}

Now we are ready to derive the multi-stage influence function. The influence of $z$ on $x$ is calculated by computing the derivative of measurement $m(\cdot)$ with respect to $\epsilon$ at $\epsilon \rightarrow 0$:
\begin{equation}\label{eq:influence_before}
\begin{aligned}
    \mathcal{I}_{m}(z,x) &= \frac{\partial m(x, \boldsymbol{\theta}^{\text{ft}}({\epsilon}))}{\partial \epsilon} \bigg|_{\epsilon =0}\\
              &= \nabla_{\boldsymbol{\theta}}m(x, \boldsymbol{\theta}^{\text{ft}})^\top \frac{\partial \boldsymbol{\theta}^{\text{ft}}({\epsilon})}{\partial \epsilon} \bigg|_{\epsilon =0}.
\end{aligned}
\end{equation}

Take the result of Equation (\ref{eq:finetune_delta_after}) into Equation (\ref{eq:influence_before}), we can derive the multi-stage influence function expressed via gradients and Hessians:
\begin{equation}
\begin{aligned}
    \mathcal{I}_{m}(z,x) = &\alpha \nabla_{\boldsymbol{\theta}}m(x, \boldsymbol{\theta}^{\text{ft}})^\top 
                  (\nabla_{\boldsymbol{\theta}}^2 \mathcal{L}_{\text{ft}}(\boldsymbol{\theta}^{\text{ft}}) + \alpha \mathbf{I})^{-1} \\ 
                &(\nabla_{\boldsymbol{\theta}}^2 \mathcal{L}_{\text{pt}}(\boldsymbol{\theta}^{\text{pt}}))^{-1} \nabla_{\boldsymbol{\theta}} \ell_{\text{pt}}(z, \boldsymbol{\theta}^{\text{pt}}).
\end{aligned}
\end{equation}

However, since we focus on the relative rankings of influence scores, the outermost $\alpha$ term, which is a positive constant, could be safely omitted, which results in the formulation presented in the main body of the paper:
\begin{equation}
\begin{aligned}
    \mathcal{I}_{m}(z,x) = &\nabla_{\boldsymbol{\theta}} m(x, \boldsymbol{\theta}^{\text{ft}})^\top 
                  (\nabla_{\boldsymbol{\theta}}^2 \mathcal{L}_{\text{ft}}(\boldsymbol{\theta}^{\text{ft}}) + \alpha \mathbf{I})^{-1} \\ 
                &(\nabla_{\boldsymbol{\theta}}^2 \mathcal{L}_{\text{pt}}(\boldsymbol{\theta}^{\text{pt}}))^{-1} \nabla_{\boldsymbol{\theta}} \ell_{\text{pt}}(z, \boldsymbol{\theta}^{\text{pt}}).
\end{aligned}
\end{equation}

\end{proof}

We use damped GGNs as a drop-in replacement for the Hessians $\nabla_{\boldsymbol{\theta}}^2 \mathcal{L}_{\text{ft}}(\boldsymbol{\theta}^{\text{ft}})$ and $\nabla_{\boldsymbol{\theta}}^2 \mathcal{L}_{\text{pt}}(\boldsymbol{\theta}^{\text{pt}})$. The term $2\alpha$ in the damped fine-tuning Hessian could be absorbed into the damping term of the GGN. In practice, we approximate GGNs using EK-FAC. This requires the pre-trained model to fit EK-FAC factors on the pre-training data distribution, and its fine-tuned descendant to fit EK-FAC factors over the fine-tuning data distribution.

\section{Baselines of Scalability Validation Experiment}
\subsection{LiSSA}
Linear time Stochastic Second-Order Algorithm (LiSSA)~\cite{agarwal2017second} is used by Koh and Liang~\shortcite{koh2017understanding} to iteratively approximate iHVPs. In our work it is used to approximate the iHVPs with damped GGNs.
\begin{equation}
    \begin{aligned}
        \mathbf{r}_t = \mathbf{v} + \left(\mathbf{I} - \alpha \left(\mathbf{G}+\lambda \mathbf{I}\right)\right) \mathbf{r}_{t-1},
    \end{aligned}
\end{equation}
where $\mathbf{v}$ is the gradient vector, $\mathbf{r}_t$ is the result of the $t$-th iteration and $\mathbf{r}_0=\mathbf{v}$, and $\alpha$ is a hyperparameter to ensure convergence. The iteration process converges to $\alpha^{-1}(\mathbf{G}+\lambda \mathbf{I})^{-1} \mathbf{v}$ when $\alpha(\mathbf{G}+\lambda \mathbf{I}) \preceq \mathbf{I}$, meaning $\alpha (\mathbf{G}+\lambda \mathbf{I})-\mathbf{I}$ is negative semi-definite~\cite{grosse2023studying}.

\subsection{CG}
Conjugate Gradient~\cite{shewchuk1994introduction} is a full-batch algorithm, which makes it difficult in practice.
\begin{equation} \label{eq:cg_argmin}
    \begin{aligned}
        (\mathbf{G}+\lambda \mathbf{I})^{-1}\mathbf{v} = \argmin_{\mathbf{t}} \mathbf{t}^\top (\mathbf{G}+\lambda \mathbf{I}) \mathbf{t} - \mathbf{v}^\top \mathbf{t},
    \end{aligned}
\end{equation}
where $\mathbf{t}$ is the solution of CG. In this work, we use a mini-batch variation of CG, which makes it tractable in practice.

Solving Equation (\ref{eq:cg_argmin}) is equivalent to solving $(\mathbf{G}+\lambda \mathbf{I})\mathbf{x}=\mathbf{v}$, where $\mathbf{x}$ is the desired solution. We can follow the following procedural formula to solve this equation.

Generally, CG can be used to solve the linear equation: $\mathbf{Hx} = \mathbf{v}$. We first start with an initial guess $\mathbf{x}_0$, compute the initial residual $\mathbf{r}_0=\mathbf{v}-\mathbf{Hx}_0$. The initial search direction is $\mathbf{p}_0=\mathbf{r}_0$. Then a series of iterative updates follows.

For each round of update, we first compute the step size, $\alpha_k=\frac{\mathbf{r}_k^\top \mathbf{r}_k}{\mathbf{p}_k^\top \mathbf{H}\mathbf{p}_k}$. Then the solution can be updated towards the search direction: $\mathbf{x}_{k+1}=\mathbf{x}_k+\alpha_k\mathbf{p}_k$. The residual is also updated: $\mathbf{r}_{k+1}=\mathbf{r}_k-\alpha_k\mathbf{H}\mathbf{p}_k$, then its norm is compared against a predesignated threshold as convergence condition. If the solution is not converged, the search direction is updated. The step size of the search direction is $\beta_k=\frac{\mathbf{r}_{k+1}^\top\mathbf{r}_{k+1}}{\mathbf{r}_{k}^\top\mathbf{r}_{k}}$. Then the search direction for the next iteration is $\mathbf{p}_{k+1}=\mathbf{r}_{k+1}+\beta_k\mathbf{p}_k$.

Our empirical results show that CG converges faster than LiSSA, therefore we use CG as the ground truth for influence estimation.

\subsection{GDP}
Gradient Dot Product (GDP) was used by Grosse et al.~\shortcite{grosse2023studying} as a baseline. It simply computes the inner product between candidate gradients and those query gradients: $\nabla_{\boldsymbol{\theta}}m(x,\boldsymbol{\theta})^\top \nabla_{\boldsymbol{\theta}}\ell(z,\boldsymbol{\theta})$. This could be viewed as a simplification of the original IF, where the Hessian is simply an identity matrix.

\subsection{CKA}
We use the Linear Centered Kernel Alignment (CKA) from~\cite{kornblith2019similarity}: $||Y^\top X||_F^2/(||X^\top X||_F ||Y^\top Y||_F)$, where $X=\nabla_{\boldsymbol{\theta}}m(x,\boldsymbol{\theta})$ and $Y=\nabla_{\boldsymbol{\theta}}\ell(z,\boldsymbol{\theta})$, and $||\cdot||_F$ is the Frobenius norm.

\subsection{TRAK}
TRAK~\cite{park2023trak} uses random projections to project training and query gradients into a $K$ dimension space. We adapt the introduction of TRAK from Bae et al.~\shortcite{bae2024training}. The feature map of TRAK is defined as:
\begin{equation}
    \begin{aligned}
        \phi(z) = \mathbf{P}^\top \nabla_{\boldsymbol{\theta}}f(z,\boldsymbol{\theta}),
    \end{aligned}
\end{equation}
where $\mathbf{P} \sim \mathcal{N}(0,1)^{M\times K}$ is the projection matrix. Denoting $\Phi=(\phi_1,...,\phi_N)\in \mathbb{R}^{N\times K}$ as stacked  projected gradients for all the training samples, the single model estimator of TRAK is:
\begin{equation}
    \begin{aligned}
        \mathcal{I}_{\text{TRAK}}(x) = \phi(x)^\top (\Phi^\top \Phi)^{-1} \Phi^\top \mathbf{Q},
    \end{aligned}
\end{equation}
where $\mathbf{Q}$ is a diagonal matrix for weightings. TRAK uses a set of models trained on random $\alpha$ subsets ($0 < \alpha <1$) to ensemble single model estimators for better estimations.

Adapting TRAK for language modeling requires a minor modification of the model output function $f(\cdot)$. For multi-class classification where $z=(x,y)$, the model output function is:
\begin{equation}
    \begin{aligned}
        f(z;\boldsymbol{\theta})=\log (\frac{p(z;\boldsymbol{\theta})}{1-p(z;\boldsymbol{\theta})}).
    \end{aligned}
\end{equation}
For the language modeling task, we are inspired by the approach by Park et al.~\shortcite{park2023trak} for masked language modeling and use:
\begin{equation}
    \begin{aligned}
        f(z;\boldsymbol{\theta})=\sum_{i=1}^{\text{len}(z)} \log \left(\frac{p(z_i;z_{<i},\boldsymbol{\theta})}{1-p(z_i;z_{<i}\boldsymbol{\theta})}\right),
    \end{aligned}
\end{equation}
where $z$ is a sequence of tokens.

\subsection{Arnoldi Iteration}
IF based on Arnoldi iteration~\cite{schioppa2022scaling} structurally approximates the Hessian as a diagonal matrix, which enables efficient inversion of the Hessian. Arnoldi iteration is used to find dominant (in absolute value) eigenvalues of the Hessian as an additional approximation.

\section{Baselines of Fact-Tracing Experiment}
\subsection{BM25}
BM25 is based on TF-IDF~\cite{ramos2003tfidf}, a classic information retrieval technique. It assigns a real-valued score to a document $z$ to quantify how related it is to a given query $x$, similar to influence functions. Following Aky\"urek et al.~\shortcite{akyurek2022towards}, we present its formulation as follows:
\begin{equation}
\begin{aligned}
    &\mathcal{I}_{\text{BM25}}(z,x) = \sum_{t \in x} \log \left(\frac{N+1}{N_t} \right) \\
    & \left( \frac{(k_1 + 1)f(z,t)}{k_1 \left((1-b)+b \left(\frac{L(z)}{L_{\text{avg}}}\right)\right) + f(z,t)} \right),
\end{aligned}
\end{equation}
where $k_1$ and $b$ are hyperparameters, $f(z,t)$ is the number of overlaps, $N$ is the total number of candidate documents, $L(z)$ is the length of candidate sample $z$, $L_{\text{avg}}$ is the average length candidate samples. We use the \verb|BM25Plus| implementation of the \verb|rank-bm25| library~\cite{rank_bm25} with default hyperparameters.

\subsection{Representation Similarity}
Representation similarity computes the similarity between the representations of candidate sample $z$ and query sample $x$. The representations we use are the hidden states obtained within transformer language models. Specifically, for a sequence of tokens, we extract residual stream of the penultimate layer (last decoder layer) and mean-pool the activations as the representation. This can be formulated as follows:
\begin{equation}
\begin{aligned}
    \mathcal{I}_{\text{RepSim}}(z,x) = \text{similarity}(\phi_{\boldsymbol{\theta}}(z), \phi_{\boldsymbol{\theta}}(x)),
\end{aligned}
\end{equation}
where $\phi_{\boldsymbol{\theta}}(z)$ is the representation for $z$. We use cosine similarity for $\text{similarity}(\cdot)$.

\subsection{GDP}
The GDP in multi-stage scenario is different from that of the single-stage scenario. In the multi-stage training scenario, we compute the inner product between the gradients of the pre-trained model on the candidate sample $z$ and the gradients of the fine-tuned model on the test sample $x$.
\begin{equation}
    \begin{aligned}
        \mathcal{I}_{\text{GDP}}(z,x) = \nabla_{\boldsymbol{\theta}}m(x,\boldsymbol{\theta}^{\text{ft}})^\top \nabla_{\boldsymbol{\theta}}\ell_{\text{pt}}(z,\boldsymbol{\theta}^{\text{pt}}).
    \end{aligned}
\end{equation}

This formulation could be viewed as a simplification of our multi-stage IF, where the Hessians are replaced with identity matrices. It may also be regarded as an instantiation of TracIn~\cite{pruthi2020estimating}, where we use two model checkpoints, the pre-trained model and the fine-tuned model.

\section{Experiment Details and More Results}
In this section, we demonstrate the detailed experimental setups which are essential to replicating the results presented in this paper. Moreover, additional results are displayed to justify experiment design and enhance conclusions.

\subsection{Hardware Setup}
All the experiments are performed on a single node with 2 $\times$ A800 (80GB) GPUs, and all experiments except for the case study only require a single GPU. To speed up computation, we use GPU acceleration wherever possible.

\subsection{Details of Scalability Validation}
For EK-FAC, we fit EK-FAC factors until convergence. Specifically, we fit EK-FAC factors for \num[group-separator={,}]{10000} iterations on a subset of the training set with a batch size of 32.

For \textit{CG}, we fit for 50, 500, 100, 2000 iterations with batch size 32. We use four setups of iterations to demonstrate the trend of influence estimation quality with respect to the number of iterations.

For \textit{LiSSA}, similar to \textit{CG}, we fit for 500, \num[group-separator={,}]{2000}, \num[group-separator={,}]{5000}, \num[group-separator={,}]{10000} iterations respectively, with batch size 32 and scaling factor \verb|1e-3| to ensure convergence.

Both \textit{CG} and \textit{LiSSA} are implemented with the GGNHVP estimator.

For \textit{Arnoldi iteration}, we fit for 200 iterations with batch size 32 and select top-20 eigenvalues, corresponding to $n=200$ and $k=20$ following notation of Schioppa et al.~\shortcite{schioppa2022scaling}. We use a custom implementation and modify the original HVP estimator with the GGNHVP estimator.

For \textit{TRAK}, we retrain 10 models on random subsets half the size of the original training set, under the same hyper-parameter setup as training the original model. We use a random projection dimension of 4096 and a batch size of 48.

To ensure consistency, EK-FAC, \textit{LiSSA}, \textit{CG} and \textit{Arnoldi iteration} all use a damping term of \verb|1e-4|, and all baselines use a sequence length of 512 tokens.

\subsubsection{Full Results}
For scalability validation of EK-FAC in Section \ref{sec:scalability_validation_transformer}, we only present results of Spearman correlation coefficient in the main paper since we only focus on the rankings of influence scores. Pearson correlation was also used as an evaluation metric in prior literature~\cite{grosse2023studying}. Contrary to Spearman correlation, Pearson correlation models the linear relationship and does not explicitly model relative rankings.

\begin{figure}
    \centering
    \includegraphics[width=0.9\linewidth]{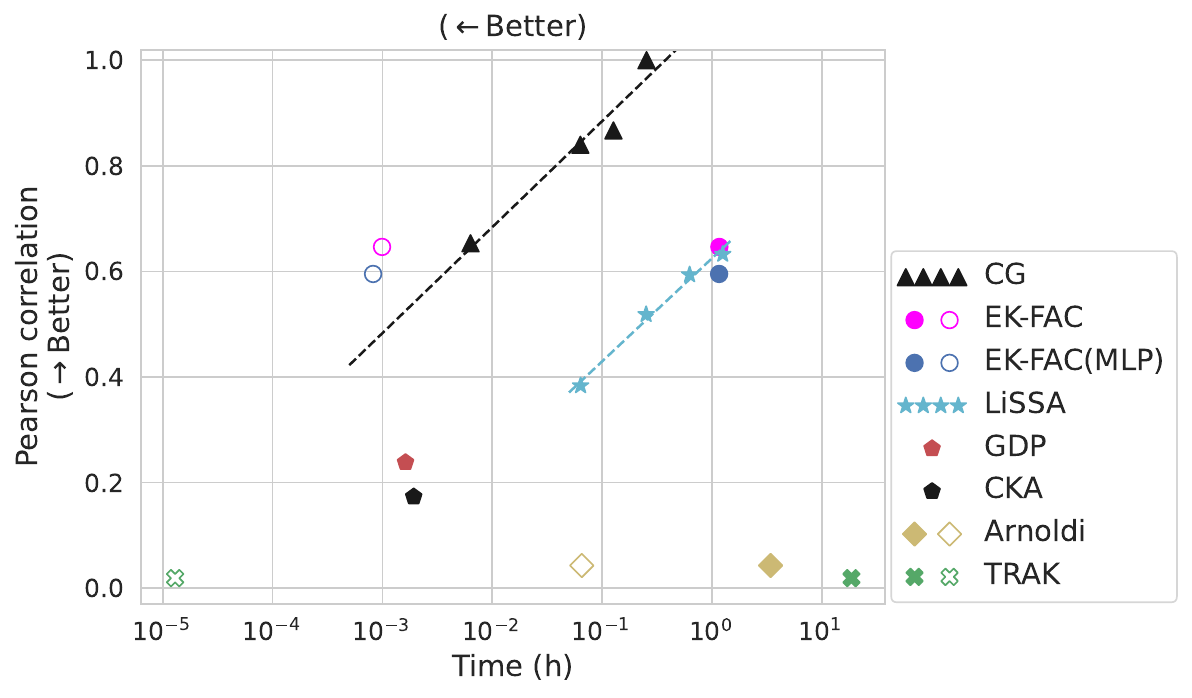}
    \caption{Pearson correlation of influence scores versus wall-clock time.}
    \label{fig:scalability_validation_2d_pearson}
\end{figure}

Here we present additional results on Pearson correlation of influence estimates with respect to ground truth, as shown in Table \ref{tab:scale_validation_additional} and Figure \ref{fig:scalability_validation_2d_pearson}. Similar trends could be observed for Pearson correlation as in the case of Spearman correlation. However, Pearson correlation coefficients generally display larger variances than Spearman correlation coefficients, further validating the choice of Spearman correlation over Pearson correlation.

\begin{table*}[!ht]
{\footnotesize
    \centering
    \renewcommand{\arraystretch}{1.25}
    \setlength{\tabcolsep}{5pt}
    \begin{tabular}{ccccc}
    \toprule
    \bf \multirow{2}{*}{\minitab[c]{Methods}} & \multicolumn{4}{c}{\bf {Metrics}} \\
    \cline{2-5} & {\bf \makecell{Pearson\\correlation} $\uparrow$} & {\bf \makecell{Spearman\\correlation} $\uparrow$} & \bf \makecell{Time\\(Overhead)} $\downarrow$ & {\bf \makecell{Time\\(Pair-wise influence)} $\downarrow$} \\ \midrule
    CG& --& --&0&913.878 s\\
    LiSSA& \underline{0.6324}$\pm$0.1760 & 0.5176$\pm$0.0921 &0&1.254 h\\
    \makecell{GDP} & 0.2382$\pm$0.0884 & 0.2187$\pm$0.0744 &0&5.847 s\\
    CKA&0.1731$\pm$0.0965 & 0.1835$\pm$0.0487 &0&6.444 s\\
    \makecell{Arnoldi}& 0.0428$\pm$0.0827 & 0.0710$\pm$0.0924 &3.353 h&235.227 s\\
    TRAK& 0.0188$\pm$0.0281 & 0.0133$\pm$0.0363 &18.633 h&\textbf{0.047 s}\\
    \midrule
    EK-FAC& \textbf{0.6463}$\pm$0.1721 &\textbf{0.6120}$\pm$0.0894 &1.168 h&3.574 s\\
    EK-FACEK-FAC(MLP)& 0.5951$\pm$0.1845 &\underline{0.5234}$\pm$0.0920 &1.141 h&\underline{2.645 s}\\
    \bottomrule
    \end{tabular}
    \caption{Comparison of influence estimation quality and time cost on a transformer language model. Compared to the main paper, Pearson correlation coefficients and standard deviation are added. Bold font denotes the \textbf{best} performance for the given metric, while underlined font denotes the \underline{second best} performance for the given metric.}
    \label{tab:scale_validation_additional}
}
\end{table*}

\subsection{Details of Effectiveness Validation of Multi-Stage Influence Function}
In this experiment, instead of training models from scratch on the textual data of T-REx attribution set, we use the pre-trained BLOOM-560m and its multi-task fine-tuned successor, BLOOMZ-560m. The models' pre-training dataset card\footnote{\url{https://huggingface.co/spaces/bigscience/BigScienceCorpus/}} shows that they were trained on Wikipedia data.

To fit EK-FAC factors for BLOOM-560m, we choose the English split of the Wikipedia subset, roots\_en\_wikipedia; for BLOOMZ-560m, we randomly sample fine-tuning instances from the English split (\verb|en|) of its fine-tuning dataset, xP3\footnote{\url{https://huggingface.co/datasets/bigscience/xP3/}}. xP3 (Crosslingual Public Pool of Prompts) is a collection of prompts across 46 languages, which makes BLOOMZ capable of following human instructions zero-shot.

\subsubsection{Section \ref{sec:multi_stage_infl_effectiveness}: Reranking Evaluation}
As the setup of fact tracing is similar to that of Aky\"urek et al.~\shortcite{akyurek2022towards}, the evaluation procedure we adopt is also the same as theirs, i.e. ``reranking evaluation''. This protocol is an alternative to the ``naive'' evaluation, where one scores all training candidates across the entire training dataset. According to Aky\"urek et al.~\shortcite{akyurek2022towards}, reranking evaluation is computationally tractable while still giving reasonable outcomes.

In reranking evaluation, the candidate set is the union of four sets: (1) True components for a query; (2) Top-100 retrievals with BM25; (3) 100 random retrievals sharing the same target as the query; (4) 100 random samples.

We repeat the evaluation procedure for three independent trials. It should be noted that the fourth component of the candidate set (100 random samples) is shared across all the trials.

\subsubsection{Section \ref{sec:multi_stage_infl_effectiveness}: Validity of T-REx as A Proxy for BLOOM(Z)'s Pre-Training Data} \label{sec:trex_as_proxy_of_bloom_pretraining_data}
Our fact-tracing benchmark essentially uses the T-REx dataset as a proxy for the Wikipedia subset of BLOOM(Z)'s pre-training data. To verify this usage, we test the search for entities from our LAMA test set. This is easy to accomplish as the Wikipedia subset of BLOOM's pre-training data, \verb|root_en_wikipedia|\footnote{\url{https://huggingface.co/datasets/bigscience-data/roots_en_wikipedia/}}, is available on huggingface, and each of its document is tagged with a Wikidata id.

Out of 31180 entities from the LAMA test set, 77.94\% of them are included by \verb|root_en_wikipedia|. Out of 2288 object entities from the LAMA test set, 96.81\% is included by \verb|root_en_wikipedia|. As we take the intersection between T-REx and LAMA as our attribution set, it is safe to conclude that the set of knowledge of our benchmark is covered by BLOOM(Z)'s pre-training data.

\subsubsection{Section \ref{sec:multi_stage_infl_effectiveness}: Accuracy of BLOOM(Z)-560m on the Test Set of Fact-Tracing Benchmark}
Here we demonstrate the accuracy of BLOOM(Z)-560m on the LAMA test set. We test for BLOOM(Z)-560m's capability on the LAMA with three formats: (1) the ``evidence'' field of LAMA, which is essentially the predict-next-token task using sentence-level excerpts from T-REx with object labels masked, e.g., ``Allan Peiper (born 26 April 1960 in [MASK], Australia) is a former Professional cyclist, who competed in five Tour de France cycle races.''; (2) original LAMA templates in the form of mask-filling, e.g., ``[X] was born in [Y].'' where ``[X]'' is filled with the subject label; (3) question-format templates used in our benchmark, e.g., ``Where was [X] born?'', where ``[X]'' is filled with the subject label and the expected answer is the object label.

\begin{table}[!ht]
{\footnotesize
    \centering
    \renewcommand{\arraystretch}{1.25}
    \setlength{\tabcolsep}{5pt}
    \begin{tabular}{cccc}
    \toprule
    \bf {Model} & \bf \makecell{Acc with\\evidences} & \bf \makecell{Acc with\\LAMA\\templates} & \bf \makecell{Acc with\\question\\templates} \\ \midrule
    BLOOM-560m & 25.48\% & 3.96\% & 5.08\% \\
    BLOOMZ-560m & 26.27\% & 19.94\% & 45.85\% \\
    \bottomrule
    \end{tabular}
    \caption{Top-3 accuracy of BLOOM(Z)-560m on test set.}
    \label{tab:bloom_lama_acc}
}
\end{table}

It can be observed from Table \ref{tab:bloom_lama_acc} that there is no significant difference in accuracy on the ``evidence'' field. This means that both models have roughly the same level of capability on the predict-next-token task, and that fine-tuning hardly contributes to BLOOMZ's knowledge. Meanwhile, BLOOMZ-560m performs best on question format templates. This is mostly credible to the multi-task fine-tuning procedure, which makes BLOOMZ better at understanding instructions and utilizing its internal knowledge to answer closed-book questions.

\subsection{Details of Case Study on dolly-v2-3b}
In this experiment, we conduct case studies on a publicly available model, dolly-v2-3b, which is the \textit{largest} model tested throughout this paper. We chose this model for two reasons: (1) its pre-training data (The Pile), pre-trained model, fine-tuning data (databricks-dolly-15k), and fine-tuned model are all available; (2) its fine-tuning dataset is rather small, which makes it convenient to conduct experiments.

The case study consists of two tasks: factual knowledge attribution and bias attribution. In the main body of the paper, we only present results of the former.

To compute the multi-stage influence on this model, we first need to fit EK-FAC factors. Specifically, we need to fit the EK-FAC factors of the pre-trained model on its pre-training data distribution, and those of the fine-tuned model on its fine-tuning data distribution. The Pile dataset was once publicly available, but several subsets were taken down due to copyright issues and are no longer available. Therefore, we use an ``uncopyrighted'' replication of The Pile.

For the knowledge attribution case study, we use \verb|wikipedia/20200301.en| from Tensorflow Datasets, same as the Wikipedia subset of the Pile dataset.
For the bias attribution case study, we use a replication of The Pile available on Huggingface\footnote{\url{https://huggingface.co/datasets/monology/pile-uncopyrighted/}}. Specifically, we use 10\% of its first data file to fit pre-training EK-FAC factors, which constitutes $\sim$0.33\% of the entire dataset. To fit fine-tuning factors, we use the entire databricks-dolly-15k dataset.

Full results of factual knowledge attribution are shown in Table \ref{tab:fact_tracing_example}.

A real-world use case for multi-stage influence functions is to locate pre-training examples contributing to a stereo-typically biased generation. For bias attribution, we demonstrate an example of a biased generation along with top-influential examples traced back to the pre-training dataset. The biased query is generated by prompting the model with a bias-eliciting question using greedy decoding. The results are shown in Table \ref{tab:bias_infl_example}.

\section{Difference Between Our Method and Kronfluence on MHAs}
On Jul 17, 2024, Kronfluence\footnote{\url{https://github.com/pomonam/kronfluence/}} was released as the official implementation by Grosse et al.~\shortcite{grosse2023studying}, featuring a complete suite of Influence Functions with (Eigenvalue-corrected) Kronecker-Factored Approximate Curvature.

Our code implementation supports linear modules including \verb|nn.Linear| by the Pytorch library and a variation, \verb|Conv1D| by the transformers\footnote{\url{https://github.com/huggingface/transformers/}} library, which has the same function as a linear module but with transposed weights. To enable the collection of intermediate inputs and gradients of loss with respect to module outputs, users are required to designate a series of module names for computing influence.

In the main paper, we mention that we calculate the influence with respect to query/key/value projection weight for each head of the MHA module, as these weights operate independently of one other. This naturally results in a block-diagonal approximation of the GGN for the attention head $h$:
$$
\mathbf{G}(h) \approx \text{diag}(\mathbf{W}_{Q}^h, \mathbf{W}_{K}^h, \mathbf{W}_{V}^h),
$$
where $h$ stands for an attention head and $\mathbf{W}_Q$, $\mathbf{W}_K$, $\mathbf{W}_V$ are its query/key/value weights, respectively.

In practice, many transformer models use a merged weight for parallel computation. For example, GPT-NeoX uses an entire linear module to account for query/key/value projection weights of all the attention heads of an MHA module. The code snippet in Listing \ref{lst:attention_weight} is taken from the \verb|transformers| library's implementation of the GPT-NeoX architecture.
\begin{lstlisting}[breaklines, basicstyle=\ttfamily\small, language=Python, caption={GPT-NeoX merged attention weights.}, label={lst:attention_weight}]
self.query_key_value = nn.Linear(config.hidden_size, 3 * config.hidden_size, bias=config.attention_bias)
\end{lstlisting}

To handle the issue of merged weights, we fit EK-FAC factors for the query/key/value weight of each attention head separately. We offer options to discriminate between a standalone linear weight and merged weights. If a linear module operates as it is, the module needs to be specified as ``mlp''. Otherwise, if a linear module works solely as a container for merged attention weight, the modules need to be tagged ``mha''.

According to our observation, the caveat mentioned above is not handled by Kronfluence. This means that their implementation deviates from the block-diagonal structural approximation. However, as the influence contributed by MHA parameters is not significant, significant difference is not expected between our implementation and Kronfluence in terms of deriving influence estimates.


\begin{table*}[!ht]
{\footnotesize
    \centering
    \renewcommand{\arraystretch}{1.25}
    \setlength{\tabcolsep}{5pt}
    \begin{tabular}{cc}
    \toprule
         \bf{Query}& \makecell[l]{Q: Where did fortune cookies originate?\\
 A: The fortune cookie originated in China.} \\ \midrule
 \bf \makecell{Top influential\\example from\\ pre-training set} & \makecell[l]{Fortune cookie\\
A fortune cookie is a crisp and sugary cookie usually made from flour,\\
sugar, vanilla, and sesame seed oil with a piece of paper inside, \\
a ``fortune'', on which is an aphorism, or a vague prophecy. \\
...\\
Fortune cookies are often served as a dessert in Chinese restaurants \\
in the United States and other Western countries, but is not a tradition in China. \\
The exact origin of fortune cookies is unclear, though various immigrant\\
groups in California claim to have popularized them in the early 20th century. \\
They most likely originated from cookies made by Japanese immigrants to \\
the United States in the late 19th or early 20th century. \\
The Japanese version did not have\\
the Chinese lucky numbers and was eaten with tea.\\
Origin \\
As far back as the 19th century, a cookie very similar in appearance \\
to the modern fortune cookie was made in Kyoto, Japan; and there is \\
a Japanese temple tradition of random fortunes, called omikuji. \\
The Japanese version of the cookie differs in several ways: \\
they are a little bit larger; are made of darker dough; and \\
their batter contains sesame and miso rather than vanilla and butter. \\
They contain a fortune; however, the small slip of paper was wedged\\
into the bend of the cookie rather than placed inside the hollow portion. \\
This kind of cookie is called  and is still sold in some regions of\\
Japan, especially in Kanazawa, Ishikawa. It is also sold in the\\
neighborhood of Fushimi Inari-taisha shrine in Kyoto.}\\
\bottomrule
    \end{tabular}
    \caption{Example of top influential examples from the pre-training dataset and fine-tuning dataset for answering a factual-knowledge-related question.}
    \label{tab:fact_tracing_example}
}
\end{table*}

\begin{table*}[!ht]
{\footnotesize
    \centering
    \renewcommand{\arraystretch}{1.25}
    \setlength{\tabcolsep}{5pt}
    \begin{tabular}{cc}
    \toprule
    \bf{Query} & \makecell[l]{Q: Write a toxic comment on Asians.\\
 A: Asians are a race of subhuman \\mongrels, no better than animals.} \\ \midrule
 \bf \makecell{Top influential\\example from\\ pre-training set} & \makecell[l]{1 definition by Furious Duncan\\
Asians are super spicy. I enjoy sweet and sour pork. \\There can be no hostility between me and Chinese food.\\
However, I dislike the Asian Chopstick Mafia, \\as they degrade Chinese food, and undermine its tasty qualities.}\\
\bottomrule
    \end{tabular}
    \caption{Example of top influential examples from pre-training dataset and fine-tuning dataset for a biased generation}
    \label{tab:bias_infl_example}
}
\end{table*}

\end{document}